\documentclass{article}

\PassOptionsToPackage{numbers, compress}{natbib}
 \usepackage[preprint]{neurips_2026}

\usepackage[utf8]{inputenc} 
\usepackage[T1]{fontenc}    
\usepackage{url}            
\usepackage{booktabs}       
\usepackage{multirow}
\usepackage{array}
\usepackage{amsfonts}       
\usepackage{nicefrac}       
\usepackage{microtype}      
\usepackage[table]{xcolor} 
\definecolor{MyGreen}{HTML}{009900}
\definecolor{MyBlue}{HTML}{0d3983}
\RequirePackage[colorlinks,citecolor=MyGreen]{hyperref}       

\usepackage{graphicx}
\usepackage{subcaption}
\usepackage{enumitem}
\usepackage{amsmath}
\usepackage{amsthm}
\usepackage{amssymb}
\usepackage{mathtools}
\usepackage{comment}
\usepackage[capitalize,noabbrev]{cleveref}
\usepackage{algorithm}
\usepackage{algorithmic}

\definecolor{tabbest}{RGB}{180,205,255}
\definecolor{tabsecond}{RGB}{212,226,255}
\definecolor{tabthird}{RGB}{235,242,255}
\newcommand{\bestcell}[1]{\cellcolor{tabbest}\textbf{#1}}
\newcommand{\secondcell}[1]{\cellcolor{tabsecond}#1}
\newcommand{\thirdcell}[1]{\cellcolor{tabthird}#1}

\theoremstyle{plain}

\theoremstyle{definition}

\theoremstyle{remark}

\title{How You Begin is How You Reason:\\
Driving Exploration in RLVR via Prefix-Tuned Priors}

  \author{
    Yifan Xu$^{1}$ \quad Junren Chen$^{2}$ \quad Yifan Chen$^{1}$ \\
    $^{1}$Hong Kong Baptist University \quad
    $^{2}$University of Maryland, College Park \\
  }

\begin{document}

\maketitle
\begin{abstract}
Reinforcement learning with verifiable rewards (RLVR) recently thrives in large language model (LLM) reasoning tasks.
However, the reward sparsity and the long reasoning horizon make effective exploration challenging.
In practice, this challenge manifests as the \emph{entropy collapse} phenomenon, where RLVR improves single-rollout accuracy but fails to expand coverage on successful reasoning trajectories.
Passive exploration techniques like entropy regularization tend to dismiss generation quality, resulting in noisy rollouts.
In response to this issue, we propose an Information-Maximizing Augmented eXploration (IMAX) framework to train a pool of soft prefixes that reshapes the base model's prior over reasoning trajectories.
Rather than relying on RL to incentivize exploration on top of the base model, 
each prefix acts as a trainable control knob that induces a distinct rollout distribution from the same backbone model.
To encourage discovery of diverse and task-relevant reasoning behaviors, we derive an Information Maximization (InfoMax) reward to complement the verifiable rewards for RL training.
IMAX is in general algorithm-agnostic and can be seamlessly integrated into existing RLVR pipelines.
Experiment results have shown that across three backbone scales, IMAX consistently improves reasoning performance over standard RLVR, with gains up to 11.60\% in Pass@4 and 10.57\% in Avg@4.
\end{abstract}


\section{Introduction}
\label{sec:intro}

Large language models (LLMs) have shown notable capabilities in complex reasoning tasks, including mathematical problem solving~\citep{jaech2024openai, guo2025deepseek, yang2025qwen3, comanici2025gemini}, automated theorem proving~\citep{yang2023leandojo, xin2024deepseek, chen2025seed}, multi-step planning~\citep{yao2022react, schick2023toolformer}, etc.
These advances in complex reasoning capabilities of LLMs are attributed to large-scale post-training procedures~\citep{chen2025sft, kumar2025llm, xu2025towards},
among which reinforcement learning from verifiable feedback (RLVR) has shown great potential and garnered widespread interest~\citep{guo2025deepseek, yu2025dapo}.
However, despite its empirical success, RLVR faces critical challenges.
Prior studies have reported an \emph{entropy collapse} phenomenon~\citep{cui2025entropy, yu2025dapo},
where the entropy of the LLM output distribution drops during training;
output diversity is thereby reduced.
In particular, some works observe that although the Pass@$1$ accuracy of the LLM significantly improves after RLVR, the Pass@$K$ performance may deteriorate compared to the base model~\citep{yue2025does, chen2025pass}.
These phenomena collectively suggest that current RLVR paradigm suffers from \emph{limited exploration}, which in turn influences reasoning capabilities.

Existing approaches for exploration enhancement of RLVR typically intervene through the RL objective, reward design, or sampling procedure, aiming to encourage broader exploration on top of the base model prior~\citep{gao2025navigate,zhang2025count,hu2025diver,shah2026upskill}.
However, recent studies suggest that RL is strongly shaped by the base model's distribution~\citep{wu2025invisible,yue2025does,karan2025reasoning}.
This observation points to a complementary direction: rather than only modifying the RL process itself, one can also steer the model prior that RLVR builds on.
In parallel, prompt-based adaptation has shown that learned prompts can effectively steer pretrained models by changing their generation context, thereby eliciting different behaviors from the same underlying model for a single prompt~\citep{li2021prefix,lester2021power,liu2022ptuning,qin2021learning}.
This motivates us to shift the optimization target from the LLM parameters to prompt tuning: rather than relying on RL to enhance exploration through samples drawn from the base model prior, we optimize lightweight conditioning variables that directly control the sampling distribution and open different regions of the model's reasoning space for each prompt.
Once learned, these prefixes can serve as reusable control handles for inference or subsequent policy training, producing rollouts that are both more diverse and more likely to contain high-quality reasoning paths.

\begin{figure*}
    \centering
    \includegraphics[width=\linewidth]{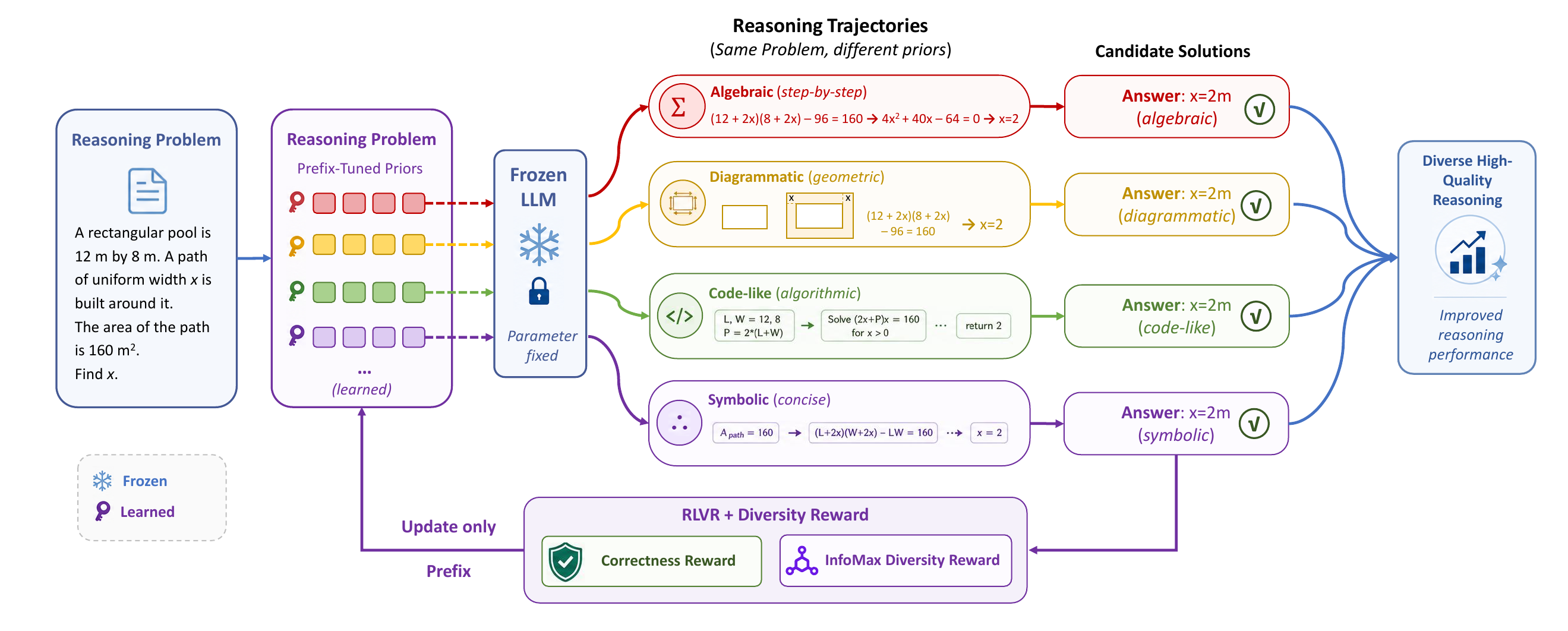}
    \caption{Overview of the IMAX framework.
    For a reasoning problem, a frozen LLM is conditioned on a learned soft-prefix pool, where different prefixes induce different reasoning priors and lead to diverse candidate trajectories, such as algebraic, diagrammatic, code-like, or symbolic solutions.
    Training updates only the prefix priors using RLVR correctness rewards and the InfoMax reward, encouraging prefixes to produce high-quality and non-redundant reasoning trajectories.}
    \label{fig:method_framework}
\end{figure*}


Therefore, we introduce an Information-Maximizing Augmented eXploration (IMAX) method on soft prompt tuning. 
The overall framework for IMAX is shown in Figure~\ref{fig:method_framework}.
We first initialize a pool of soft prompts, each soft prompt acts as a lightweight conditioning handle: when prepended to the same frozen backbone model, it shifts the rollout distribution toward a different region of the model's reasoning space.
We then train these prompts with RLVR so that the induced priors favor correct reasoning, meanwhile adding an intrinsic Information Maximization (InfoMax) reward to prevent the prompt pool from collapsing into redundant ones.
Concretely, we maximize the mutual information (MI) between the prompt identity and the generated trajectory.
The resulting objective encourages prompts to induce distinct trajectories for different soft prompts, yielding a lightweight method for improving both diversity and quality during inference. 
And to encourage the soft prompts to explore in semantically-meaningful regions, we initialize soft prompts from textual prompts that encode different reasoning strategies.
In summary, our main contributions in this paper are threefold:
\begin{enumerate}[itemsep=-0.2em, topsep=0.0em, leftmargin=1.0em]
    \item We introduce a lightweight RLVR paradigm for learning a soft prompt pool that reshapes the prior of a frozen reasoning model without updating the full policy.
    \item We propose an InfoMax reward that encourages different prompts to induce distinguishable, task-relevant reasoning trajectories, preventing collapse to redundant prefixes.
    \item We conduct extensive experiments on challenging mathematical reasoning benchmarks, verifying that the learned prompt pool improves inference diversity and performance over existing baselines.
\end{enumerate}

\section{Related Works}
\label{sec:related_works}

We briefly introduce the related topics in this section, and provide a detailed discussion in Appendix~\ref{app: related_works}.

\paragraph{Reinforcement learning from verifiable rewards.}

RLVR has become a prominent approach for improving LLM reasoning with verifiable feedback, with progress highlighted by OpenAI o1~\cite{jaech2024openai}, DeepSeek-R1~\citep{guo2025deepseek}, and Group Relative Policy Optimization (GRPO) from DeepSeekMath~\citep{shao2024deepseekmath}. These advances have sparked growing research on RLVR for
LLM reasoning~\citep{yu2025dapo, zheng2025group, tan2025gtpo}. However, RLVR still faces key
challenges, including the reported ``entropy collapse''~\citep{cui2025entropy, wang2025beyond} and
debates over whether it truly improves reasoning or mainly increases confidence within the original
distribution, potentially driving models toward determinism rather than better
exploration~\citep{yue2025does,zhu2025surprising}.

\paragraph{Exploration in RLVR.}

Exploration remains a central challenge in RLVR. Recent studies show that effective exploration requires preserving informative tokens, not simply increasing randomness~\citep{cui2025entropy,wang2025beyond,huang2025low,gai2025differential,chen2025empirical}. Existing methods address this exploration issue by modifying the optimization
objective~\citep{yu2025dapo,chen2025pass}, adding exploration or diversity
bonuses~\citep{gao2025navigate,zhang2025count,hu2025diver}, or intervening in rollouts through entropy-targeted tokens~\citep{zheng2025first,bai2026learning}.
A concurrent work~\citep{shah2026upskill} also uses mutual information to improve RLVR exploration. 
However, our method differs fundamentally in both objective and algorithm design: IMAX optimizes in the soft-prompt space instead of the full model parameter space, and we also initialize prefixes from semantically meaningful prompts that encode distinct reasoning strategies. A detailed comparison of the two methods is provided in Appendix~\ref{app: related_works}.

\paragraph{Soft prompting.}

Prompt-based adaptation includes hard prompting, soft prompting, and layer-wise prefix tuning. Prior work shows that prompts can act as lightweight control variables for frozen models, eliciting latent behaviors~\citep{qin2021learning}, transferring across tasks~\citep{vu2022spot}, or activating pretrained capabilities~\citep{petrov2024when}. Prompt pools further use prompt collections as compact task memories or input-conditioned control modules~\citep{wang2022learning,wang2022dualprompt,smith2023coda}. 
In this paper, we use classic soft prompt tuning, where learned continuous virtual tokens are prepended to each input prompt. We refer to these vectors as soft prompts or prefixes and use the two terms interchangeably throughout the paper.

\section{Preliminaries}
\label{sec:preliminaries}


We first introduce the soft-prompt parameterization, then formulate RLVR under prefix tuning, and finally present GRPO, the main RLVR algorithm used in this paper.




\subsection{Soft Prompt Tuning}
\label{sec:soft_prompt_tuning}

Soft prompt tuning adapts a frozen language model by optimizing continuous prompt embeddings.
Given an input prompt $x=(x_1,\dots,x_L)$, a soft prompt $e=[e_1,\dots,e_m]\in\mathbb{R}^{m\times d}$ consists of $m$ trainable virtual-token embeddings with the same hidden dimension $d$ as the model token embeddings.
The model conditions on the concatenated input-embedding sequence $[e_1,\dots,e_m,\mathrm{Emb}(x_1),\dots,\mathrm{Emb}(x_L)]$, where $\mathrm{Emb}(\cdot)$ denotes the input embedding table of the base model.
Thus, the soft prompt is prepended at the input-embedding layer before the Transformer blocks, while the backbone parameters remain fixed.
When initialized from a text prompt, the initialization text is tokenized and mapped through the input embedding table and the resulting vectors initialize the soft prompt and are then optimized directly as continuous parameters.

\subsection{RLVR for Prefix Tuning}

We consider RLVR for prefix tuning, where the objective is to improve reasoning performance using verifiable reward signals while updating only the soft prefixes.
The backbone LLM remains frozen, and the trainable parameters are a prefix pool $\mathcal{E}=\{e_1,\ldots,e_C\}$ with $e_z\in\mathbb{R}^{m\times d}$.
For a sampled prefix index $z$, the model generates a response from
\begin{equation}
\pi_{\mathcal{E}}(y \mid x,z)
\equiv
\pi_{\theta_0}(y\mid x,e_z)
=
\prod_{t=1}^{T}
\pi_{\theta_0}(y_t \mid x,e_z,y_{<t}),
\end{equation}

where $\theta_0$ is the parameter for the base LLM. In RLVR, the environment provides a binary reward $R(x,y) \in \{0,1\}$ for each prompt-response pair, indicating whether the final answer is correct.
The prefix-tuning RLVR objective is to maximize the expected reward over both prefix sampling and response generation:
\begin{equation}
\max_{\mathcal{E}} \;
\mathbb{E}_{x\sim p(x),\,z\sim p(z),\,y \sim \pi_{\mathcal{E}}(\cdot \mid x,z)}
\left[ R(x,y) \right].
\end{equation}




\subsection{Formulation of GRPO}

GRPO is a group-relative RL algorithm specifically designed for LLMs that avoids training a separate critic by estimating advantages through \emph{group-wise relative comparison} among sampled responses.
In our prefix-tuning setting, the group is formed within each assigned prefix.
Given an input prompt $x$ and a prefix index $z$, GRPO samples a group of $N$ responses $\{y^{(n)}\}_{n=1}^{N}$ under the same prefix-conditioned policy, $y^{(n)}\sim \pi_{\mathcal{E}_{\text{old}}}(\cdot \mid x,z)$, where $\mathcal{E}_{\text{old}}$ denotes the prefix pool used to collect trajectories.
If multiple prefix identities are assigned to the same prompt, each prompt--prefix pair $(x,z)$ defines its own GRPO group.

The responses are evaluated by the verifier to obtain rewards $\{ R(x, y^{(n)}) \}_{n=1}^{N}$, and the normalized advantage is $A^{(n)}=(R(x,y^{(n)})-\mu_R)/\sigma_R$, where $\mu_R$ and $\sigma_R$ are the group mean and standard deviation.
The clipped GRPO objective is
\begin{equation}
\begin{aligned}
    \mathcal{L}_{\text{GRPO}}(\mathcal{E})=\mathbb{E}_{x,n,t}\big[\min\big(r^{(n)}_t(\mathcal{E})A^{(n)},\tilde{A}^{(n)}_t(\mathcal{E})\big)\big],
    \quad
    r^{(n)}_t(\mathcal{E})=
    \frac{\pi_{\mathcal{E}}(y^{(n)}_t \mid x,z, y^{(n)}_{<t})}
    {\pi_{\mathcal{E}_{\text{old}}}(y^{(n)}_t \mid x,z, y^{(n)}_{<t})}.
\end{aligned}
\end{equation}
where $\tilde{A}^{(n)}_t(\mathcal{E})=\operatorname{clip}\big(r^{(n)}_t(\mathcal{E}), 1-\epsilon, 1+\epsilon\big)A^{(n)}$ is the clipped surrogate term and $\epsilon$ is a clipping hyperparameter.
In the original GRPO formulation, a KL regularizer is added to limit policy deviation. As our goal is to encourage prefix-conditioned exploration, we omit this KL term, following the practice adopted in DAPO~\citep{yu2025dapo}.

\section{Methodology}
\label{sec:methodology}

In this section, we first introduce prefix-conditioned reasoning in Section~\ref{sec:prefix_priors},
and then derive a variational lower bound of the InfoMax reward in Section~\ref{sec:variational MI} for practical implementation.
The detailed post-training procedure is stated in \Cref{sec:detail-procedure}.

\subsection{Prefix-Conditioned Reasoning}
\label{sec:prefix_priors}



We train a soft-prefix pool $\mathcal{E}=\{e_1,\dots,e_C\}$, where each prefix index $z$ selects a distinct rollout distribution $\pi_{\mathcal{E}}(y\mid x,z)$.
We initialize the prefix pool from a text prompt pool $\mathcal{T}={t_1,\dots,t_C}$ generated by GPT-5.4~\citep{singh2025openai}, with deliberate prompting on producing diverse and minimally overlapping initial prompts. The resulting default prompt pool is listed in Table~\ref{tab:default_prefix_pool} in Appendix~\ref{app:default_prefix_pool}.
Each text prompt $t_z$ is mapped through the frozen embedding layer of the backbone model and used to initialize $e_z$, after which $e_z$ is optimized as continuous parameters.
This initialization gives the prefix pool semantically meaningful and diverse starting points.

\begin{figure}[htbp]
    \centering
    \includegraphics[width=\textwidth]{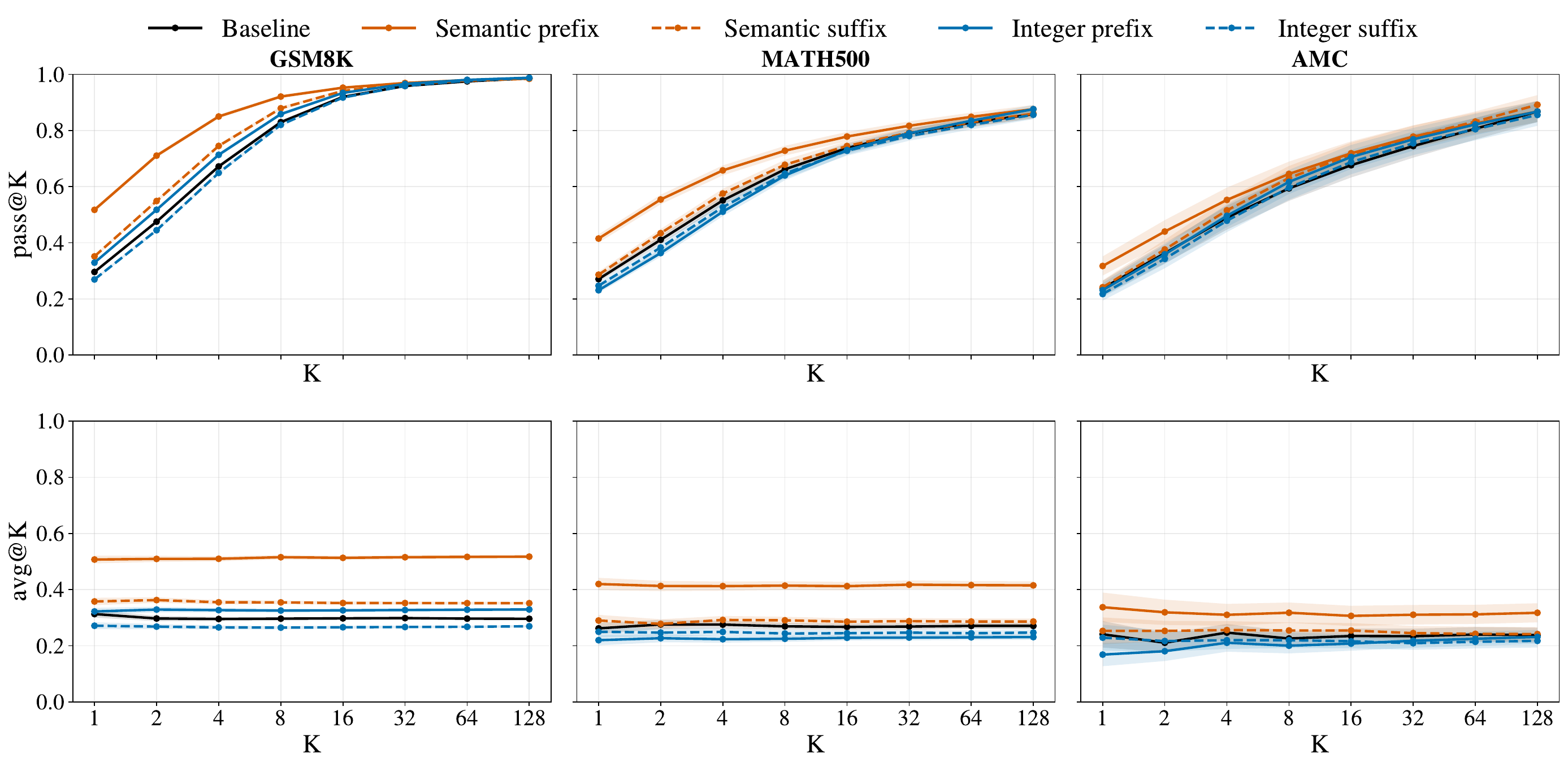}
    \caption{
    Inference-time performance of training-free prompting.
    We compare the base model with semantic and integer prompts inserted either as prefixes or suffixes across GSM8K, MATH500, and AMC on Qwen2.5-Math-1.5B.
    The top row reports Pass@K, while the bottom row reports Avg@K.
    }
    \label{fig:preliminary_exp}
\end{figure}

We first conduct an inference-time experiment on Qwen2.5-MATH-1.5B~\citep{yang2024qwen25math} to show that changing the generation condition can indeed shift the base model's inference behavior, thereby justifying prefix-level control of the sampling prior.
As shown in Figure~\ref{fig:preliminary_exp}, appending semantically meaningful prefixes yields clear Pass@$K$ gains over the base model at small to moderate $K$, which is precisely the rollout regime typically used by current group-relative RLVR algorithms.
The improvement becomes thinner as $K$ grows larger.
This effect is consistently stronger when the control is inserted as a prefix rather than a suffix, while integer labels rarely produce comparable gains.
Moreover, semantic prefix conditioning substantially improves Avg@$K$ across the three benchmarks, indicating that it raises the average quality of sampled reasoning trajectories.

To make the prefix pool useful, different prefixes should induce high-quality and distinguishable reasoning trajectories.
To enable the former, we use RL algorithm to perform online training. 
The requirement of diversity is to prevent the case that learned pool may collapse to redundant prefixes, where
$\pi_{\mathcal{E}}(y\mid x,z)\approx \pi_{\mathcal{E}}(y\mid x,z')$ for different $z$ and $z'$.
Therefore, we adopt an \emph{information-guided principle}: the prefix identity should capture salient variations in the inherent reasoning paths conditioned on the same input.
Concretely, for a fixed prompt $x$, this requires maximizing the fixed-prompt mutual information between the prefix identity $z$ and the generated response $y$
\begin{equation*}
I(Z; Y\mid X)
=
\mathbb{E}_{p(x)}\left[\mathbb{E}_{p(z,y\mid x)}
\left[
\log \frac{p(z,y\mid x)}{p(z\mid x)p(y\mid x)}
\right]\right],
\end{equation*}
thereby encouraging different soft prefixes to induce distinct and valid reasoning behaviors for the same question.


Alternatively, we may write in terms of joint and conditional entropies as
\begin{equation}
\begin{aligned}
    I(Z;Y\mid X)
    &= H(Y \mid X) - H(Y \mid X, Z)
    && \text{// forward} \\
    &= H(Z \mid X) - H(Z \mid X, Y)
    && \text{// reverse}
\end{aligned}
\label{eq:MI}
\end{equation}

where $H(\cdot)$ denotes entropy. 
In Equation~\eqref{eq:MI}, ``forward'' and ``reverse'' denote two equivalent decompositions. The forward form follows the generation direction $Z\!\to\!Y$, while the reverse form follows the inference direction $Y\!\to\!Z$.
Intuitively, the mutual information term $I(Z;Y\mid X)$ is zero if the response variable $Y$ is independent of the prefix variable $Z$ given the prompt $X$, and increases as the prefix-conditioned rollout distributions become more distinguishable.


\subsection{Variational InfoMax Reward}
\label{sec:variational MI}

To incorporate the InfoMax reward into prefix-pool training,
we consider the mutual information in the reverse form of Equation~\eqref{eq:MI} for practical computation.
A detailed discussion on this choice is deferred to Appendix~\ref{app:InfoMax_discussion}.
We first rewrite the conditional mutual information as
\begin{equation}
\begin{aligned}
I(Z;Y\mid X) &\coloneqq H(Z\mid X)-H(Z\mid Y,X)
    =\mathbb{E}_{p(x)}\big[\mathbb{E}_{p(z,y\mid x)}\big[\log p(z\mid x,y) 
    - \log p(z\mid x)\big]\big].
\end{aligned}
\end{equation}
The posterior $p(z \mid x,y)$ is generally intractable as in variational inference, and we similarly introduce a variational distribution $q_\phi(z\mid x,y)$ to obtain the variational lower bound for $p(z\mid x,y)$.
Since $z$ indexes one of $C$ prefix identities, $q_\phi$ is implemented as a $C$-way classifier that predicts which soft prefix generated the response.
In particular, now $I(Z;Y\mid X)$ reads
\small
\begin{align*}
\mathbb{E}_{p(x)}\left[\mathbb{E}_{p(z,y\mid x)}\big[\log q_\phi(z\mid x,y) - \log p(z\mid x)\big]\right]
+\mathbb{E}_{p(x)}\left[\mathbb{E}_{p(y\mid x)}\left[\mathrm{KL}\!\left(p(z\mid x,y)\,\|\,q_\phi(z\mid x,y)\right)\right]\right],
\end{align*}
\normalsize
and by dropping the last non-negative KL divergence term, we can immediately have the lower bound of $I(Z;Y\mid X)$ as
\begin{equation}
\begin{aligned}
\mathbb{E}_{p(x)}\left[\mathbb{E}_{p(z,y\mid x)}\big[\log q_\phi(z\mid x,y) - \log p(z\mid x)\big]\right].
\end{aligned}
\end{equation}
The above expectation is taken with respect to the conditional distribution $p(z,y\mid x)$.
We leverage a standard transform
to rewrite the expectation under a tractable sampling scheme, $p(z\mid x) \cdot p(y\mid x,z)$.
Here, $z$ is sampled from the prefix-identity distribution and the response $y$ is sampled from $p(\cdot\mid x,z)$, which corresponds to the frozen LLM distribution $\pi_{\mathcal{E}}(\cdot \mid x,z)$ conditioned on both the prompt and the selected soft prefix.

For the variational objective, we consider the case where the prefix prior is independent of the prompt, where $p(z\mid x)=p(z)=\mathrm{Uniform}\{1,\dots,C\}$,
which is a standard design choice also adopted in previous works~\citep{chen2016infoGAN}.
Consequently, the mutual information objective reduces to
\begin{equation}
    \begin{aligned}
        I(Z;Y\mid X) &\ge\mathbb{E}_{p(x)}\big[\mathbb{E}_{p(z,y\mid x)}\big[\log q_\phi(z\mid x,y)-\log p(z)\big]\big] \\
        &= \mathbb{E}_{p(x,z,y)}\big[\log q_\phi(z\mid x,y)\big]+H(z).
    \end{aligned}
\label{eqn:vlb-mi}
\end{equation}
Here $H(z)$ is constant across all samples, and we omit it from the objective.

We then incorporate the variational lower bound in Equation~\eqref{eqn:vlb-mi} for conditional mutual information into the RLVR objective as the InfoMax reward, which now reads
\begin{equation}
\max_{\mathcal{E}} \;
\mathbb{E}_{x \sim p(x),z \sim p(z),y \sim \pi_{\mathcal{E}}(\cdot \mid x,z)}
\Big[
R(x,y)
+
\beta\log q_\phi(z\mid x,y)
\Big].
\end{equation}
A hyperparameter $\beta$ controls the strength of the InfoMax reward compared with the verifiable reward.
Notably, this InfoMax reward is defined at the response level: $q_\phi$ predicts the prefix identity from the prompt $x$ and the complete generated response $y$, because the semantic reasoning path induced by a prefix is expressed by the trajectory as a whole rather than by isolated token-level deviations.



\subsection{Detailed Post-training Procedure}
\label{sec:detail-procedure}

We detail the adaptation of the above variational framework to our prefix-tuning procedure in this section.
The variational posterior network $q_\phi(z\mid x,y)$ encodes the prompt--response sequence with the same frozen LLM backbone and applies a $C$-way classification head to the hidden state of the last non-padding token.
Only this classification head is trained during update.

\paragraph{Alternating optimization.}
We update $q_\phi$ and the prefix pool $\mathcal{E}$ in an alternating manner for each online training batch.
This alternating procedure follows the spirit of coordinate ascent variational inference (CAVI)~\citep{blei2017variational}. With the prefix pool fixed, updating $q_\phi$ tightens the variational posterior used in the InfoMax lower bound; with $q_\phi$ fixed, updating $\mathcal{E}$ improves the prefix-conditioned rollout distribution under the current variational reward estimate.
Before updating the prefix pool, we ensure that the proxy posterior network $q_\phi$ is sufficiently updated within the batch so that the variational lower bound can be reliably estimated, as the InfoMax reward must be computed prior to the RLVR optimization step.
To compute $q_\phi(z \mid x,y)$, we perform an additional forward pass of the frozen backbone on the original prompt--response pair $(x,y)$ without the soft prefix, thereby preventing information leakage of $z$ through the prefix embedding itself.
When training the posterior network $q_\phi$, the prefix pool is fixed.
Since $q_\phi$ is lightweight, its training incurs negligible computational overhead.
After updating $q_\phi$, the prefix pool $\mathcal{E}$ is optimized with the combined verifiable reward and InfoMax reward using the RLVR algorithm.

\paragraph{Incorporation into RLVR paradigm.}
Specifically, some adaptations are required for the group-relative RL algorithms.
Let $C$ denote the number of soft prefixes and $N$ denote the number of rollouts in each GRPO group.
For each input prompt $x$, we use a stratified prefix assignment: the prompt is paired with distinct prefix identities $z\in\{1,\dots,C\}$, and the corresponding continuous prefix $e_z$ is prepended to form the prefix-conditioned input.
For each prompt--prefix pair $(x,z)$, we sample $N$ rollouts from $\pi_{\mathcal{E}}(\cdot\mid x,z)$ and treat these $N$ responses as one GRPO group, under which the group-relative baseline can be computed in a standard manner.
This stratified assignment keeps the empirical marginal over prefix identities uniform, and yields a prompt layout that is more convenient for batched parallel generation than random sampling in modern rollout engines.
The overall training algorithm of IMAX is shown in Algorithm~\ref{alg:imax} at Appendix~\ref{app:algorithm}.

\section{Experiments}
\label{sec:experiments}


\subsection{Experimental Setup}
\label{sec:exp_setup}
\paragraph{Models and baselines.}
We evaluate our method on Qwen2.5-1.5B-Instruct~\citep{yang2024qwen25}, Qwen3-4B-Base, and Qwen3-8B-Base~\citep{yang2025qwen3}.
We compare against full-policy GRPO (GRPO-full), text-prompt initialization (Text Init.), and representative RLVR baselines including GRPO (GRPO-prefix), DAPO (DAPO-prefix), DIVER (DIVER-prefix)~\citep{hu2025diver}, and MERCI (MERCI-prefix)~\citep{zhang2025merci} under the prefix-tuning setting. For the record, we adopt the exploration intrinsic rewards in DIVER and MERCI, and explicitly adapt them into our prefix-tuning setting.
We adopt the default reward weights used in the original papers.
We train the models on the MATH~\citep{hendrycks2021MATH} dataset, which consists of 7.5k rows of data.

\paragraph{Implementation details.}
We implement our algorithms based on the TRL framework~\footnote{https://github.com/huggingface/trl.},
all experiments are conducted on 1 × RTX PRO 6000-96GB GPU.
we set the number of prefixes assigned to each prompt to be $C=2$,
and set the GRPO group size to $N=4$ rollouts per prompt--prefix pair.
For optimization, we use AdamW with a learning rate of $5 \times 10^{-4}$.
The weight for the InfoMax reward is set to be 0.01.
More implementation details are reported in Appendix~\ref{app:experiment_setup}.

\paragraph{Evaluation Details.}
For evaluation, we use lighteval~\footnote{https://github.com/huggingface/lighteval.} to test the algorithms on mathematical reasoning benchmarks including {MATH-500}~\citep{lightman2023let}, {GSM8K}~\citep{cobbe2021GSM8K}, Minerva Math~\citep{lewkowycz2022solving}, and {AMC}~\citep{hendrycks2021MATH}, and to show that our tuned prefixes do not degrade performance on general tasks, we also evaluate on an instruction following task--IFEval~\citep{zhou2023instruction}.
For prefix-tuned methods, evaluation uses the same stratified prefix assignment paradigm as in training: each prompt is paired with distinct $C$ learned prefixes and the same number of rollouts $N_\text{eval}$ is generated for each prompt--prefix pair. We set $C=2$, and $N_\text{eval}=2$, yielding rollouts number $K=4$ for each original prompt in evaluation.
We report Pass@$4$ and Avg@$4$ as the evaluation metrics.

\subsection{Main Results}
\label{sec:main_exp}

\begin{table*}[t]
\centering
\caption{Main results on reasoning and instruction-following benchmarks. We report percentages (\%) for readability, the full table with standard errors is provided in Appendix~\ref{app:additional_results}. MATH is the abbreviation for MATH-500. Darker shading marks stronger results within each metric block, while unshaded Base and GRPO-full rows are shown as reference baselines.}
\label{tab:main_results}
\footnotesize
\setlength{\tabcolsep}{3pt}
\resizebox{\textwidth}{!}{
\begin{tabular}{ll|ccccc|ccccc}
\toprule
& & \multicolumn{5}{c|}{\textbf{Pass@4}} & \multicolumn{5}{c}{\textbf{Avg@4}} \\
\cmidrule(lr){3-7}\cmidrule(lr){8-12}
\textbf{Model} & \textbf{Method} & \textbf{GSM8K} & \textbf{MATH} & \textbf{Minerva} & \textbf{AMC} & \textbf{IFEval} & \textbf{GSM8K} & \textbf{MATH} & \textbf{Minerva} & \textbf{AMC} & \textbf{IFEval} \\
\midrule
\multirow{10}{*}{Qwen2.5-1.5B}
& Base & 78.92 & 49.60 & 13.97 & 68.22 & 52.31 & 53.43 & 32.85 & 6.99 & 35.28 & 37.75 \\
& GRPO-full & 84.84 & \textbf{60.20} & 18.01 & 46.04 & 50.28 & \textbf{74.09} & \textbf{47.55} & \textbf{10.94} & 22.79 & 35.26 \\
\cmidrule(lr){2-12}
& Text Init. & 80.29 & 50.20 & 16.18 & 68.31 & \thirdcell{54.34} & 56.92 & 34.35 & 8.82 & 34.70 & \secondcell{38.08} \\
& DIVER-prefix & 80.29 & 56.20 & \thirdcell{16.54} & 67.72 & \bestcell{54.90} & 59.76 & 37.05 & 7.72 & 33.28 & \thirdcell{38.03} \\
& MERCI-prefix & 82.11 & \thirdcell{59.00} & 16.18 & 68.31 & \secondcell{54.53} & 63.34 & \thirdcell{40.90} & 8.64 & \thirdcell{35.05} & 37.43 \\
& GRPO-prefix & \secondcell{85.75} & 49.40 & 15.44 & \bestcell{69.22} & 53.60 & 66.89 & 34.80 & 8.27 & \secondcell{35.38} & 37.11 \\
& GRPO+IMAX & \secondcell{85.75} & \bestcell{60.00} & \bestcell{18.38} & \thirdcell{68.06} & 53.05 & \thirdcell{67.40} & \secondcell{42.30} & \secondcell{9.47} & 34.45 & 36.78 \\
& DAPO-prefix & \bestcell{86.28} & 56.40 & \secondcell{17.28} & \secondcell{68.97} & 51.94 & \bestcell{69.50} & \thirdcell{41.60} & \thirdcell{9.19} & \bestcell{35.53} & 36.74 \\
& DAPO+IMAX & \thirdcell{84.99} & \secondcell{59.40} & 15.81 & \thirdcell{68.39} & 53.79 & \secondcell{67.63} & \bestcell{42.50} & \bestcell{9.83} & 34.99 & \bestcell{38.12} \\
\midrule
\multirow{9}{*}{Qwen3-4B}
& Base & 90.60 & 52.00 & 22.43 & 54.13 & 48.80 & 59.70 & 30.00 & 10.57 & 22.83 & 26.94 \\
& GRPO-full & 91.81 & \textbf{69.20} & 25.37 & 16.76 & 43.25 & \textbf{86.64} & \textbf{59.80} & \textbf{20.13} & 9.78 & 29.25 \\
\cmidrule(lr){2-12}
& Text Init. & 88.55 & 51.20 & \secondcell{24.26} & 67.39 & 50.65 & 53.49 & 27.50 & 10.02 & 35.22 & 28.97 \\
& DIVER-prefix & 91.43 & 57.80 & 23.53 & 66.97 & \thirdcell{52.31} & 66.49 & 31.75 & 9.93 & 35.38 & 30.41 \\
& MERCI-prefix & 90.83 & \thirdcell{60.80} & 20.96 & 64.80 & 51.20 & 62.51 & 30.75 & 9.56 & 34.92 & \thirdcell{31.47} \\
& GRPO-prefix & \thirdcell{93.48} & 57.60 & 21.69 & \secondcell{70.64} & 51.39 & \thirdcell{71.97} & \secondcell{43.45} & \secondcell{13.33} & \bestcell{39.64} & \thirdcell{31.93} \\
& GRPO+IMAX & \bestcell{94.47} & \bestcell{69.20} & \thirdcell{23.90} & \bestcell{71.06} & 51.57 & \bestcell{76.14} & \bestcell{50.45} & \thirdcell{11.86} & \thirdcell{38.07} & 30.64 \\
& DAPO-prefix & 93.18 & 56.60 & \bestcell{25.74} & \thirdcell{69.06} & \bestcell{56.19} & \thirdcell{72.02} & \thirdcell{41.65} & \bestcell{14.52} & \secondcell{38.32} & \bestcell{35.95} \\
& DAPO+IMAX & \secondcell{93.78} & \secondcell{65.80} & \thirdcell{23.90} & 64.97 & \secondcell{55.08} & \secondcell{72.93} & \thirdcell{42.70} & 10.85 & 35.78 & \secondcell{33.13} \\
\midrule
\multirow{9}{*}{Qwen3-8B}
& Base & 88.10 & 58.60 & 18.01 & 41.12 & 48.06 & 55.44 & 33.65 & 8.73 & 18.58 & 27.26 \\
& GRPO-full & 91.58 & \textbf{74.20} & \textbf{29.41} & \textbf{74.23} & 47.87 & \textbf{76.95} & \textbf{65.30} & \textbf{22.33} & \textbf{47.46} & 31.24 \\
\cmidrule(lr){2-12}
& Text Init. & 90.30 & 59.80 & 20.96 & 62.39 & 52.50 & 56.16 & 33.60 & 8.36 & 31.15 & 30.87 \\
& DIVER-prefix & 90.75 & 63.00 & \thirdcell{21.69} & 62.89 & \bestcell{55.27} & 57.15 & 36.65 & \thirdcell{9.56} & \thirdcell{29.69} & \secondcell{32.76} \\
& MERCI-prefix & \secondcell{93.18} & \thirdcell{66.40} & 18.38 & \secondcell{66.47} & \secondcell{54.71} & \thirdcell{59.59} & 37.95 & 8.64 & \thirdcell{31.86} & \secondcell{32.76} \\
& GRPO-prefix & 91.78 & \thirdcell{66.40} & 19.12 & 64.14 & \secondcell{54.71} & \secondcell{60.90} & \thirdcell{40.20} & 9.19 & \bestcell{33.36} & \bestcell{33.60} \\
& GRPO+IMAX & \bestcell{94.01} & \bestcell{67.20} & 20.96 & \bestcell{67.06} & 51.57 & \bestcell{64.18} & \secondcell{41.00} & \secondcell{10.02} & \secondcell{32.42} & 30.59 \\
& DAPO-prefix & 90.07 & 63.60 & \secondcell{22.43} & \thirdcell{65.47} & \thirdcell{53.79} & 52.12 & 34.60 & \thirdcell{9.56} & \thirdcell{31.86} & \thirdcell{31.05} \\
& DAPO+IMAX & \thirdcell{92.72} & \secondcell{67.00} & \bestcell{23.16} & 63.55 & 51.76 & \thirdcell{60.20} & \bestcell{41.10} & \bestcell{11.40} & \thirdcell{32.24} & 30.45 \\
\bottomrule
\end{tabular}
}
\end{table*}


\textbf{RLVR for prefix tuning is an effective adaptation strategy.}
Table~\ref{tab:main_results} shows that applying RLVR to the prefix pool improves both Pass@4 and Avg@4 over the frozen base model on most reasoning benchmarks. On Qwen2.5-1.5B, for example, GRPO-prefix raises AMC Pass@4 from 46.04\% to 69.22\% and GSM8K Avg@4 from 53.43\% to 66.89\%, while DAPO-prefix improves GSM8K Pass@4 from 78.92\% to 86.28\% and MATH Avg@4 from 32.85\% to 41.60\%. These gains are obtained without updating the backbone parameters, showing that optimizing only the prefix pool can already produce meaningful reasoning adaptation. Prefix-tuned RLVR is also competitive with full-policy GRPO in Pass@4 on several benchmarks, while causing no significant degradation on IFEval.

\textbf{IMAX further improves RLVR for prefix-tuning.}
Compared with GRPO and DAPO under the same prefix-tuning setting, IMAX consistently brings additional gains on reasoning benchmarks: on Qwen3-4B-Base, GRPO+IMAX improves over GRPO on GSM8K, MATH-500, and AMC, and DAPO+IMAX improves over DAPO on MATH-500. On Qwen3-8B-Base, DAPO+IMAX further improves over DAPO on GSM8K, MATH-500, and Minerva. IMAX also compares favorably with diversity-oriented baselines such as DIVER and MERCI, especially on Qwen3-4B-Base where GRPO+IMAX achieves the strongest Pass@4 on GSM8K, MATH, and AMC among prefix-tuning methods. These results suggest that the InfoMax reward does more than add generic diversity: on top of RLVR, IMAX improves \emph{effective exploration} which remains useful for verifiable reasoning.

\subsection{Ablation Studies}
\label{sec:ablation}
\textbf{Effectiveness of RL training.}
Table~\ref{tab:ablation_results} compares supervised prefix tuning with IMAX-based prefix RLVR, and further ablates the number of prefixes assigned to each prompt.
The number of group rollouts $N$ is kept the same across runs, which suggest that more prefixes yield more rollouts.
Compared with SFT, RL-based prefix tuning gives substantially stronger results on most reasoning benchmarks, especially on Avg@4.
This confirms that online update is important for turning soft prefixes into useful reasoning controllers, rather than merely fitting them to supervised demonstrations.

\textbf{Scaling of the prefix pool.}
Table~\ref{tab:ablation_results} also shows an overall increasing tendency as more prefixes are assigned to each prompt.
This trend supports the role of the prefix pool: different prefixes expose different regions of the frozen model's reasoning distribution, and RLVR can exploit this expanded set of prefix-conditioned rollouts.
The full ablation table with standard errors is reported in Table~\ref{tab:full_ablation}.

\begin{table*}[t]
\centering
\caption{Ablation results of training algorithm and prefix number on Qwen2.5-1.5B-Instruct.}
\label{tab:ablation_results}
\footnotesize
\setlength{\tabcolsep}{3pt}
\resizebox{0.92\textwidth}{!}{
\begin{tabular}{l|ccccc|ccccc}
\toprule
& \multicolumn{5}{c|}{\textbf{Pass@4}} & \multicolumn{5}{c}{\textbf{Avg@4}} \\
\cmidrule(lr){2-6}\cmidrule(lr){7-11}
\textbf{Run} & \textbf{GSM8K} & \textbf{MATH} & \textbf{Minerva} & \textbf{AMC} & \textbf{IFEval} & \textbf{GSM8K} & \textbf{MATH} & \textbf{Minerva} & \textbf{AMC} & \textbf{IFEval} \\
\midrule
SFT & 83.40 & 54.80 & 14.70 & \secondcell{68.80} & \bestcell{54.70} & 58.40 & 28.50 & 6.00 & 29.90 & 35.80 \\
IMAX-1 prefix & \thirdcell{83.60} & \secondcell{58.60} & \secondcell{18.80} & \bestcell{69.30} & \secondcell{54.50} & \thirdcell{64.30} & \thirdcell{40.30} & \thirdcell{9.00} & \bestcell{34.90} & \bestcell{39.10} \\
IMAX-2 prefixes & \secondcell{85.75} & \bestcell{60.00} & \thirdcell{18.38} & \thirdcell{68.06} & \thirdcell{53.05} & \secondcell{67.40} & \bestcell{42.30} & \secondcell{9.47} & \secondcell{34.45} & \thirdcell{36.78} \\
IMAX-4 prefixes & \bestcell{86.60} & \thirdcell{58.40} & \bestcell{19.50} & 67.60 & \bestcell{54.70} & \bestcell{68.20} & \secondcell{41.60} & \bestcell{9.50} & \bestcell{34.90} & \secondcell{37.60} \\
\bottomrule
\end{tabular}
}
\end{table*}

\begin{figure*}[t]
\centering
\includegraphics[width=\textwidth]{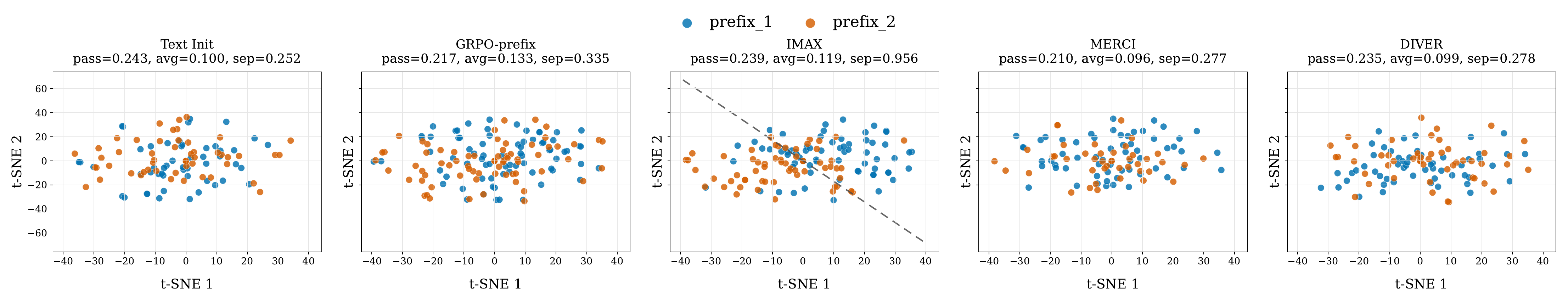}
\caption{t-SNE visualization of response embeddings for the two trained prefixes on Minerva Math for Qwen3-4B-Base under different prefix-tuning methods. The separation score is the distance between the two soft-prompt centroids normalized by their average within-cluster spread. We provide complete embedding figures in Figure~\ref{fig:complete_embedding_analysis} at Appendix~\ref{app:additional_results}.}
\label{fig:minerva_tsne}
\end{figure*}

\subsection{Further Analysis}
We first list the two initialized prefixes used in this analysis, so that the subsequent comparisons of reasoning components can be interpreted with respect to their intended semantic roles.

\label{sec:analysis}

\begin{center}
\begin{minipage}{0.96\linewidth}
\small
\setlength{\fboxsep}{6pt}
\colorbox{gray!8}{
\begin{minipage}{0.96\linewidth}
\textbf{Prefix 1.} Introduce variables explicitly, write down the governing equations or identities,
and solve through exact symbolic manipulation. Prefer precise derivation over heuristic guessing.

\vspace{2pt}
\textbf{Prefix 2.} Partition the problem into exhaustive, mutually exclusive cases. Solve each case
carefully, rule out impossible ones, and combine the valid conclusions.
\end{minipage}}
\end{minipage}
\end{center}

\begin{figure*}[t]
\centering
\begin{subfigure}[t]{0.32\textwidth}
    \centering
    \includegraphics[width=\linewidth]{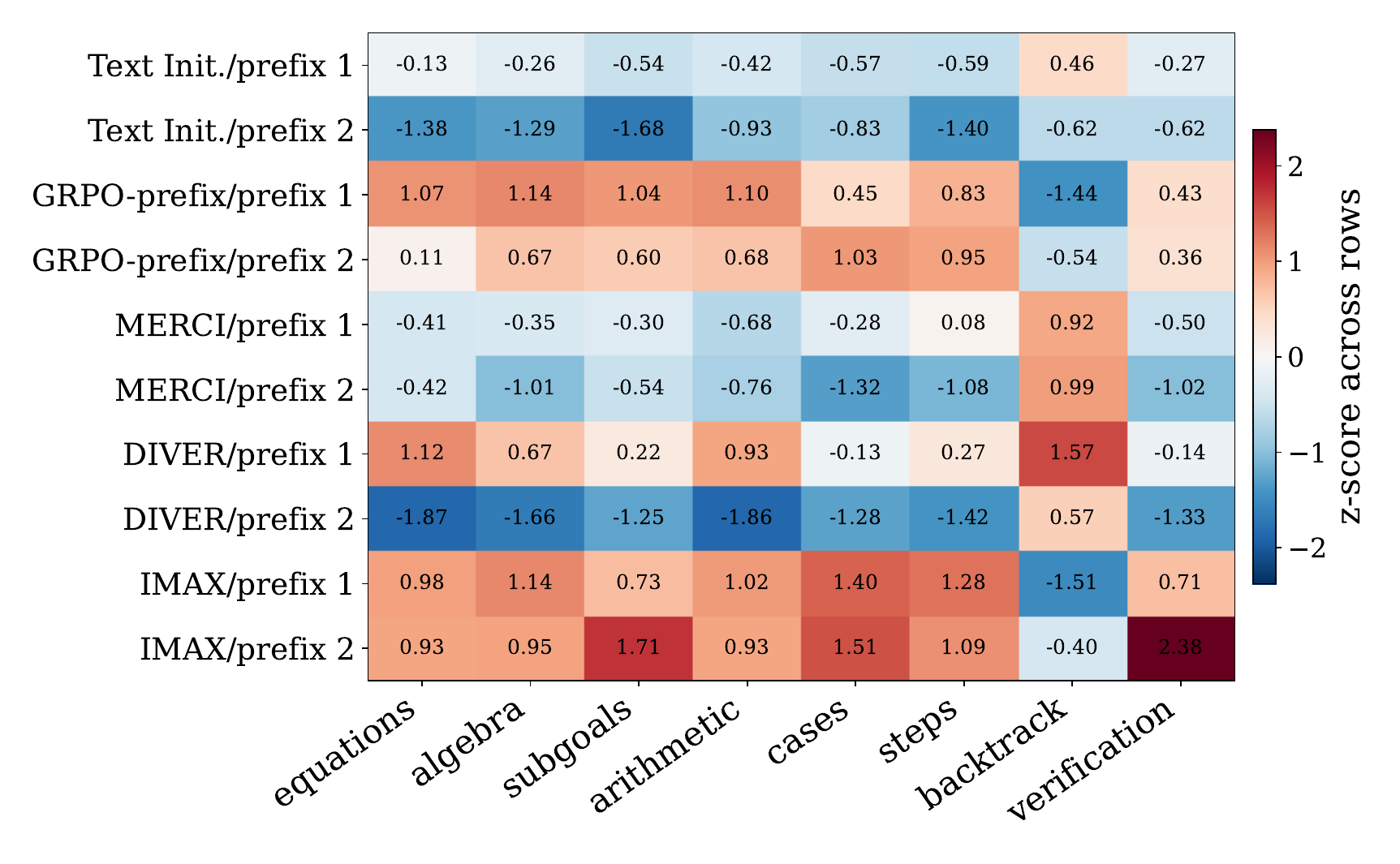}
    \caption{Reasoning component heatmap.}
    \label{fig:math500_analysis_triptych_heatmap}
\end{subfigure}\hfill
\begin{subfigure}[t]{0.32\textwidth}
    \centering
    \includegraphics[width=\linewidth]{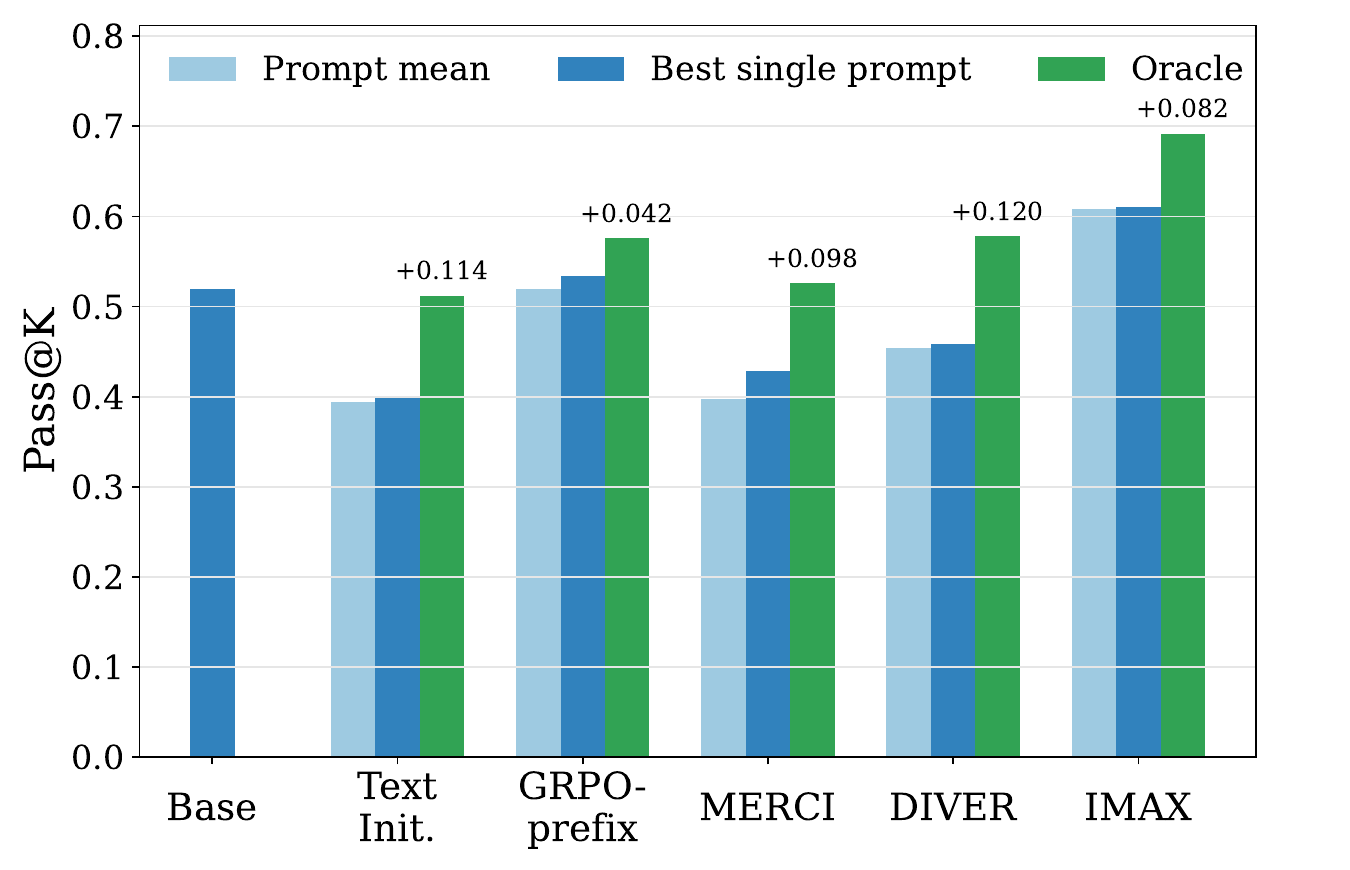}
    \caption{Prefix correctness.}
    \label{fig:math500_analysis_triptych_correctness}
\end{subfigure}\hfill
\begin{subfigure}[t]{0.32\textwidth}
    \centering
    \includegraphics[width=\linewidth]{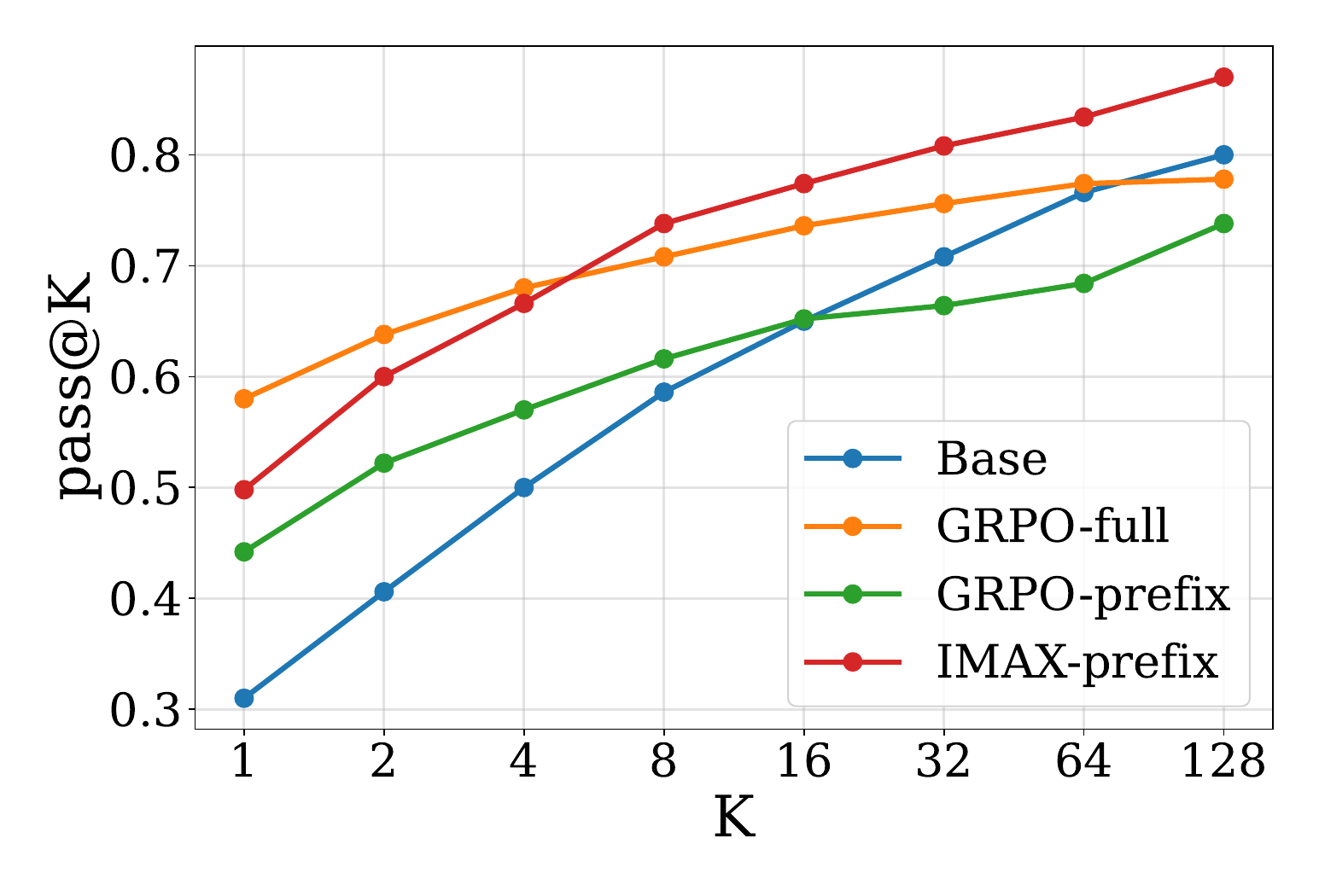}
    \caption{Pass@$K$ comparison.}
    \label{fig:math500_analysis_triptych_passk}
\end{subfigure}
\caption{Analysis of prefix-conditioned reasoning behavior on MATH500 for Qwen3-4B-Base. (a): the heatmap visualizes the counts of different components appeared in the reasoning trajectories of Prefix 1 and Prefix 2 for the prefix-tuning methods. (b): the correctness bar shows the mean, best and combined (oracle) pass rate of the two prefixes, higher oracle suggests that the two prefixes solve partially overlapping but non-identical subsets of problems. (c): the Pass@$K$ comparison shows how prefix-conditioned sampling affects coverage as the rollout budget increases. We present complete analysis features in Figure~\ref{fig:complete_heatmap_analysis} and Figure~\ref{fig:complete_correctness_analysis} at Appendix~\ref{app:additional_results}.}
\label{fig:math500_analysis_triptych}
\end{figure*}

\textbf{IMAX discovers prefix-specific reasoning patterns.}
We next examine whether IMAX induces distinct reasoning behaviors within the prefix group.
Figure~\ref{fig:minerva_tsne} shows that IMAX yields more separated answer-embedding clusters across prefixes, indicating stronger intra-group diversity. This diversity is also semantically aligned with the prefix designs: in \Cref{fig:math500_analysis_triptych_heatmap}, Prefix 1 induces more equation, algebra, and arithmetic operations, while Prefix 2 induces more subgoals, cases, backtracking, and verification. The contrast is strongest for subgoals and verification, where IMAX recovers the intended prefix-specific pattern more clearly than competing methods. These results suggest that IMAX encourages structured reasoning diversity rather than surface-level variation.

\textbf{IMAX enhances effective exploration.}
\Cref{fig:math500_analysis_triptych_correctness} shows that the two prefixes cover complementary successful regions as the combined oracle beats single best-performing prefix, while GRPO-prefix improves correctness but reduces prefix diversity. In contrast, IMAX preserves complementary coverage while maintaining correctness. Together with Figure~\ref{fig:minerva_tsne}, this suggests that IMAX induces structured reasoning diversity rather than merely separating outputs. Moreover, \Cref{fig:math500_analysis_triptych_passk} of Pass@$K$ performance shows that although GRPO-full is strongest at small $K$, IMAX surpasses it when $K>4$ and remains above the base model, indicating improved successful-region coverage.

\textbf{Runtime efficiency analysis.} 
We report the runtime for each method in Figure~\ref{fig:runtime_analysis} at Appendix~\ref{app:additional_results}, due to limited space in the main context.
The additional runtime of IMAX over standard prefix-tuned RLVR baselines mainly comes from the extra forward pass used to obtain the prompt--response representation for training the posterior model $q_\phi$. Even with this overhead, IMAX remains substantially cheaper than full-parameter GRPO (measured in GPU-hours). IMAX requires about $3.0\times$--$6.0\times$ less training cost across model scales. This advantage becomes more important as the backbone grows, because prefix-tuned methods only incur increased rollout cost from the base model, whereas full-parameter tuning also scales the backward and optimizer costs over all model parameters.

\section{Conclusion}
\label{sec:conclusion}

In this work, we study effective exploration in RLVR for LLM reasoning from the perspective of prefix-tuning.
We propose Information-Maximizing Augmented eXploration (IMAX), a lightweight post-training framework that optimizes a pool of soft prompts to reshape the model's conditional generation distribution.
To prevent the learned prefixes from collapsing into redundant controls, we introduce a response-level InfoMax reward, and optimize the posterior estimator as well as the prefix pool in an alternating procedure.
Experiments on mathematical reasoning benchmarks show that IMAX improves both reasoning quality and prefix-conditioned diversity over standard prefix-tuned RLVR baselines. 
This opens a direction for treating soft prompts as reusable control variables for both inference-time reasoning and future exploration-aware RLVR training.

\paragraph{Limitations}

Our method is mainly designed for reasoning tasks: reasoning-incentivized prefixes improve
mathematical reasoning benchmarks, but bring limited gains on general instruction-following tasks. Its effectiveness also depends on prefix initialization, and our current text-prompt initialization requires manual design and may require retuning across domains. Finally, the learned soft prefixes must be appended during inference, introducing a small but persistent prompt-side overhead.




\bibliographystyle{plainnat}
\bibliography{ref}

\newpage
\appendix
\onecolumn
\begin{center}
{\Large\bfseries Appendix}
\end{center}







\section{Complete Related Works}
\label{app: related_works}

We first review RLVR algorithms for LLM reasoning and then discuss the exploration challenge in RLVR. We next introduce soft prompt tuning, which forms the basis of our setting. Finally, we provide a detailed comparison with concurrent work.

\paragraph{Reinforcement learning from verifiable rewards.}

As large language models (LLMs) show strong capabilities across tasks, improving their reasoning ability has become critical.
Among existing approaches, reinforcement learning from verifiable rewards (RLVR) has attracted growing attention for enhancing LLM reasoning with verifiable feedback.
Early reinforcement learning demonstrated its potential for reasoning tasks with the release of the OpenAI o1 model~\cite{jaech2024openai}.
Subsequently, the first open-source reasoning model, DeepSeek-R1~\citep{guo2025deepseek}, was released, achieving comparable performance.
Earlier, DeepSeekMath~\citep{shao2024deepseekmath} introduced group relative policy optimization (GRPO), which has since become a widely used RLVR algorithm for LLM post-training.
These developments have sparked a wave of research on RLVR for LLM reasoning~\citep{yu2025dapo, zheng2025group, tan2025gtpo}.
Despite the rapid progress of RLVR, this paradigm still faces challenges.
Several studies have reported an ``entropy collapse'' phenomenon of RLVR~\citep{cui2025entropy, wang2025beyond}.
Moreover, debates on whether RLVR genuinely enhances reasoning or merely increases confidence in the original distribution support the concern that RLVR may drive models toward determinism rather than improved exploration~\citep{yue2025does,zhu2025surprising}.



\paragraph{Exploration in RLVR.}

Exploration is a central difficulty in RLVR because verifiable rewards are sparse and are typically observed only after complete reasoning trajectories are sampled.
Recent analyses show that RLVR can rapidly concentrate probability mass and that effective learning often depends on preserving informative high-entropy or low-probability decision points, rather than uniformly increasing randomness~\citep{cui2025entropy,wang2025beyond,huang2025low,gai2025differential,chen2025empirical}.
One line of work addresses this issue by modifying the optimization objective, such as DAPO~\citep{yu2025dapo} and Pass@K-oriented training that rebalance exploitation and exploration~\citep{chen2025pass}.
Another line introduces explicit exploration bonuses or diversity incentives through inrinsic rewards~\citep{gao2025navigate,zhang2025count,hu2025diver}.
Other methods intervene in the rollout process, for example by targeting entropy-eliciting tokens~\citep{zheng2025first} or injecting parameter-space noise~\citep{bai2026learning}.

\paragraph{Soft prompting.}

Prompt-based adaptation spans several mechanisms: hard prompt search optimizes discrete tokens in the vocabulary~\citep{shin2020autoprompt,deng2022rlprompt}, soft prompt tuning learns continuous virtual tokens prepended to the input embedding sequence~\citep{lester2021power,liu2022ptuning}, and prefix-tuning injects trainable prefix key-value states into each Transformer layer for generation~\citep{li2021prefix}.
These methods support the view that prompts act as lightweight control variables for frozen models: mixtures of soft prompts can query different latent behaviors from the same LM~\citep{qin2021learning}, source prompts can transfer across tasks~\citep{vu2022spot}, and theoretical analyses suggest that prompting and prefix-tuning often work by eliciting capabilities already present in the pretrained model rather than rewriting the model's internal computation~\citep{petrov2024when}.
Prompt-pool methods further show that collections of prompts can serve as compact task memories or input-conditioned control modules~\citep{wang2022learning,wang2022dualprompt,smith2023coda}.
However, existing RL-based prompt optimization mainly targets discrete prompt search~\citep{deng2022rlprompt}, and diversity-aware prompt pools have been studied primarily for task selection or continual learning rather than for RLVR reasoning rollouts.
Our work instead optimizes a pool of soft prompts with verifiable rewards and an InfoMax reward so that different prompts induce distinct, high-quality reasoning trajectories.


\section{Discussion on the InfoMax Reward}
\label{app:InfoMax_discussion}

In the InfoMax reward design, we choose the reverse form of the mutual information to derive the variational lower bound.
For the forward form, consider the following decomposition of the mutual information objective:

\begin{equation*}
\begin{aligned}
I(y;z\mid x)
&=
H(y\mid x) - H(y\mid x,z) \\
&=
\mathbb{E}_{p(x)}\!\left[
\mathbb{E}_{p(y\mid x)}\big[-\log p(y\mid x)\big]
\right]
-
\mathbb{E}_{p(x)}\!\left[
\mathbb{E}_{p(z,y\mid x)}\big[-\log p(y\mid x,z)\big]
\right] \\
&=
\mathbb{E}_{p(x)}\!\left[
\mathbb{E}_{p(z,y\mid x)}\big[\log p(y\mid x,z)\big]
-
\mathbb{E}_{p(y\mid x)}\big[\log p(y\mid x)\big]
\right],
\end{aligned}
\end{equation*}
\begin{equation*}
\begin{aligned}
\text{where}\quad p(z,y\mid x) &= p(z\mid x)\,p(y\mid x,z), p(y\mid x) = \mathbb{E}_{p(z\mid x)}\!\left[p(y\mid x,z)\right].
\end{aligned}
\end{equation*}

In LLMs, we know that $p(y\mid x,z)$ is the forward log probability of response $y$ given question $x$ and latent prefix $z$ as condition.
However, the second term is an intractable marginalization term over the latent prefix $z$,

\begin{equation*}
    \mathbb{E}_{p(y\mid x)}\big[\log p(y\mid x)\big]=\int_z\mathbb{E}_{p(y\mid x)}\big[\log p(z)p(y\mid x,z)\big]dz.
\end{equation*}

\textbf{A biased contrastive target.} It is worth noting that under the latent-variable formulation, this term does not correspond to a single LLM forward pass on $(x, y)$.
Rather, it implicitly involves marginalizing over the latent prefix, evaluating how well the output can be explained across different prefix-conditioned models induced by the prior.
This is because the model defines $p(y\mid x)$ only after marginalizing over the latent prefix.
Concretely, the LLM induces a family of prefix-conditioned distributions, and the likelihood of $y$ is obtained by aggregating their contributions under the prior rather than evaluating a single prefix instance.
As a result, this term cannot be computed by a single forward pass on $(x,y)$ with a fixed or absent latent variable.
If this term is instead approximated by a single forward pass with a fixed or ignored latent prefix, the resulting objective no longer corresponds to the true mutual information.
In this case, it effectively becomes a \emph{contrastive target}: it measures the gap between the likelihood under a \emph{single} prefix-conditioned model $p(y\mid x,z)$ (or an implicitly fixed $z$) and the likelihood under the prefix-marginalized model $p(y\mid x)$.
Intuitively, this contrasts ``with $z$'' versus ``without $z$'', rather than capturing the dependence between $z$ and $y$ averaged over the latent space.
Optimizing such an objective tends to encourage the policy to exploit a particular $z$ (or ignore $z$ altogether), leading to biased gradients, latent-space collapse, and reduced diversity of prefix-coupled reasoning trajectories.

\textbf{Intractable marginalization.}
Approximating this term requires sampling multiple latent prefixes per question to account for marginalization, which substantially increases the sampling cost.
This burden is further amplified in group-relative RLVR algorithms that already rely on multiple rollouts per prompt.
Consequently, we do not adopt this form of mutual information when deriving our InfoMax reward.

\section{Additional Experiment Results}
\label{app:additional_results}

This appendix provides the complete experimental evidence omitted from the main text for space. We first report full benchmark tables with standard errors, then provide the complete ablation table, followed by additional visual analyses of prefix-conditioned behavior and runtime cost.

\subsection{Main results}

Tables~\ref{tab:full_qwen25},~\ref{tab:full_qwen3}, and~\ref{tab:full_qwen38} report the complete Pass@4 and Avg@4 results of Qwen2.5-1.5B-Instruct, Qwen3-4B-Base, and Qwen3-8B-Base, respectively. All values are percentages, and standard errors are reported for both metrics. These tables expand Table~\ref{tab:main_results} by including the full benchmark set, including GPQA, AIME24, and AIME25.

For $N$ evaluation examples, we compute the standard error over per-example scores $x_i$:
\begin{equation}
\bar{x}=\frac{1}{N}\sum_{i=1}^{N}x_i,
\qquad
\mathrm{stderr}(x)=
\sqrt{\frac{1}{N(N-1)}\sum_{i=1}^{N}(x_i-\bar{x})^2}.
\end{equation}
For Pass@$K$, $x_i=\mathbf{1}[\exists j\le K:c_{ij}=1]$; for Avg@$K$, $x_i=\frac{1}{K_i}\sum_{j=1}^{K_i}c_{ij}$, where $c_{ij}\in\{0,1\}$ indicates whether rollout $j$ for example $i$ is correct and $K_i=\min(K,\text{number of rollouts for example }i)$. The reported Pass@$K$ and Avg@$K$ values are the means $\frac{1}{N}\sum_i x_i$ under the corresponding per-example definition.

\begin{table*}[t]
\centering
\caption{Full results for Qwen2.5-1.5B-Instruct.}
\label{tab:full_qwen25}
\scriptsize
\resizebox{\textwidth}{!}{
\begin{tabular}{l|cccccccc}
\toprule
\multicolumn{9}{c}{\textbf{Pass@4}} \\
\midrule
\textbf{Method} & \textbf{GSM8K} & \textbf{MATH} & \textbf{Minerva} & \textbf{GPQA} & \textbf{AMC} & \textbf{AIME24} & \textbf{AIME25} & \textbf{IFEval} \\
\midrule
Base & $78.92{\pm}1.33$ & $49.60{\pm}2.24$ & $13.97{\pm}2.11$ & $62.63{\pm}3.45$ & $68.22{\pm}1.35$ & $13.33{\pm}6.22$ & $0.00{\pm}0.00$ & $52.31{\pm}2.15$ \\
GRPO-full & $84.84{\pm}0.99$ & $60.20{\pm}2.19$ & $18.01{\pm}2.33$ & $60.10{\pm}3.49$ & $46.04{\pm}1.44$ & $13.33{\pm}6.31$ & $6.67{\pm}4.63$ & $50.28{\pm}0.00$ \\
\cmidrule(lr){1-9}
Text Init. & $80.29{\pm}1.30$ & $50.20{\pm}2.24$ & $16.18{\pm}2.24$ & \thirdcell{$64.65{\pm}3.41$} & $68.31{\pm}1.35$ & \secondcell{$6.67{\pm}4.63$} & \bestcell{$6.67{\pm}4.63$} & \thirdcell{$54.34{\pm}2.14$} \\
DIVER-prefix & $80.29{\pm}1.10$ & $56.20{\pm}2.22$ & \thirdcell{$16.54{\pm}2.26$} & \bestcell{$65.66{\pm}3.38$} & $67.72{\pm}1.35$ & \secondcell{$6.67{\pm}4.63$} & \secondcell{$3.33{\pm}3.33$} & \bestcell{$54.90{\pm}2.14$} \\
MERCI-prefix & $82.11{\pm}1.06$ & \thirdcell{$59.00{\pm}2.20$} & $16.18{\pm}2.24$ & $57.58{\pm}3.52$ & $68.31{\pm}1.34$ & \thirdcell{$3.33{\pm}3.33$} & \thirdcell{$0.00{\pm}0.00$} & \secondcell{$54.53{\pm}2.14$} \\
GRPO-prefix & \secondcell{$85.75{\pm}1.17$} & $49.40{\pm}2.24$ & $15.44{\pm}2.20$ & $60.61{\pm}3.48$ & \bestcell{$69.22{\pm}1.34$} & $0.00{\pm}0.00$ & \bestcell{$6.67{\pm}4.63$} & $53.60{\pm}2.15$ \\
GRPO-prefix+IMAX & \secondcell{$85.75{\pm}1.17$} & \bestcell{$60.00{\pm}2.19$} & \bestcell{$18.38{\pm}2.35$} & $62.12{\pm}3.46$ & $68.06{\pm}1.35$ & $0.00{\pm}0.00$ & \secondcell{$3.33{\pm}3.33$} & $53.05{\pm}2.15$ \\
DAPO-prefix & \bestcell{$86.28{\pm}0.95$} & $56.40{\pm}2.22$ & \secondcell{$17.28{\pm}2.30$} & \secondcell{$65.15{\pm}3.39$} & \secondcell{$68.97{\pm}1.34$} & \bestcell{$10.00{\pm}5.57$} & \thirdcell{$0.00{\pm}0.00$} & $51.94{\pm}0.00$ \\
DAPO-prefix+IMAX & \thirdcell{$84.99{\pm}0.98$} & \secondcell{$59.40{\pm}2.20$} & $15.81{\pm}2.22$ & $60.10{\pm}3.49$ & \thirdcell{$68.39{\pm}1.34$} & \secondcell{$6.67{\pm}4.63$} & \bestcell{$6.67{\pm}4.63$} & $53.79{\pm}0.00$ \\
\midrule
\multicolumn{9}{c}{\textbf{Avg@4}} \\
\midrule
\textbf{Method} & \textbf{GSM8K} & \textbf{MATH} & \textbf{Minerva} & \textbf{GPQA} & \textbf{AMC} & \textbf{AIME24} & \textbf{AIME25} & \textbf{IFEval} \\
\midrule
Base & $53.43{\pm}1.23$ & $32.85{\pm}1.76$ & $6.99{\pm}1.22$ & $26.77{\pm}1.93$ & $35.28{\pm}0.96$ & $3.33{\pm}1.56$ & $0.00{\pm}0.00$ & $37.75{\pm}1.81$ \\
GRPO-full & $74.09{\pm}1.05$ & $47.55{\pm}2.00$ & $10.94{\pm}1.61$ & $28.41{\pm}2.12$ & $22.79{\pm}0.89$ & $5.83{\pm}3.10$ & $2.50{\pm}1.84$ & $35.26{\pm}0.00$ \\
\cmidrule(lr){1-9}
Text Init. & $56.92{\pm}1.23$ & $34.35{\pm}1.81$ & $8.82{\pm}1.40$ & \bestcell{$28.03{\pm}1.92$} & $34.70{\pm}0.94$ & \thirdcell{$1.67{\pm}1.16$} & \bestcell{$1.67{\pm}1.16$} & \secondcell{$38.08{\pm}1.77$} \\
DIVER-prefix & $59.76{\pm}1.08$ & $37.05{\pm}1.74$ & $7.72{\pm}1.24$ & \secondcell{$27.53{\pm}1.87$} & $33.28{\pm}0.91$ & \secondcell{$2.50{\pm}1.84$} & \secondcell{$0.83{\pm}0.83$} & \thirdcell{$38.03{\pm}1.76$} \\
MERCI-prefix & $63.34{\pm}1.07$ & $40.90{\pm}1.83$ & $8.64{\pm}1.35$ & $23.61{\pm}1.83$ & \thirdcell{$35.05{\pm}0.95$} & \thirdcell{$1.67{\pm}1.67$} & \thirdcell{$0.00{\pm}0.00$} & $37.43{\pm}1.75$ \\
GRPO-prefix & $66.89{\pm}1.22$ & $34.80{\pm}1.84$ & $8.27{\pm}1.35$ & $27.15{\pm}1.94$ & \secondcell{$35.38{\pm}0.95$} & $0.00{\pm}0.00$ & \bestcell{$1.67{\pm}1.16$} & $37.11{\pm}1.75$ \\
GRPO-prefix+IMAX & \thirdcell{$67.40{\pm}1.22$} & \secondcell{$42.30{\pm}1.86$} & \secondcell{$9.47{\pm}1.43$} & $26.77{\pm}1.97$ & $34.45{\pm}0.95$ & $0.00{\pm}0.00$ & \secondcell{$0.83{\pm}0.83$} & $36.78{\pm}1.75$ \\
DAPO-prefix & \bestcell{$69.50{\pm}1.02$} & \thirdcell{$41.60{\pm}1.91$} & \thirdcell{$9.19{\pm}1.41$} & \thirdcell{$27.40{\pm}1.83$} & \bestcell{$35.53{\pm}0.95$} & \bestcell{$4.17{\pm}2.42$} & \thirdcell{$0.00{\pm}0.00$} & $36.74{\pm}0.00$ \\
DAPO-prefix+IMAX & \secondcell{$67.63{\pm}1.03$} & \bestcell{$42.50{\pm}1.88$} & \bestcell{$9.83{\pm}1.57$} & $25.38{\pm}1.91$ & $34.99{\pm}0.95$ & \thirdcell{$1.67{\pm}1.16$} & \bestcell{$1.67{\pm}1.16$} & \bestcell{$38.12{\pm}0.00$} \\
\bottomrule
\end{tabular}
}
\end{table*}

\begin{table*}[t]
\centering
\caption{Full results for Qwen3-4B-Base.}
\label{tab:full_qwen3}
\scriptsize
\resizebox{\textwidth}{!}{
\begin{tabular}{l|cccccccc}
\toprule
\multicolumn{9}{c}{\textbf{Pass@4}} \\
\midrule
\textbf{Method} & \textbf{GSM8K} & \textbf{MATH} & \textbf{Minerva} & \textbf{GPQA} & \textbf{AMC} & \textbf{AIME24} & \textbf{AIME25} & \textbf{IFEval} \\
\midrule
Base & $90.60{\pm}0.80$ & $52.00{\pm}2.24$ & $22.43{\pm}2.53$ & $68.69{\pm}3.30$ & $54.13{\pm}1.44$ & $10.00{\pm}5.57$ & $3.33{\pm}3.33$ & $48.80{\pm}2.15$ \\
GRPO-full & $91.81{\pm}0.76$ & $69.20{\pm}2.07$ & $25.37{\pm}2.64$ & $52.53{\pm}3.56$ & $16.76{\pm}1.08$ & $10.00{\pm}5.57$ & $13.33{\pm}6.31$ & $43.25{\pm}0.00$ \\
\cmidrule(lr){1-9}
Text Init. & $88.55{\pm}0.88$ & $51.20{\pm}2.24$ & \secondcell{$24.26{\pm}2.60$} & \thirdcell{$58.08{\pm}3.52$} & $67.39{\pm}1.35$ & \secondcell{$16.67{\pm}6.92$} & $3.33{\pm}3.33$ & $50.65{\pm}2.15$ \\
DIVER-prefix & $91.43{\pm}0.77$ & $57.80{\pm}2.21$ & $23.53{\pm}2.58$ & \secondcell{$59.60{\pm}3.50$} & $66.97{\pm}1.36$ & $6.67{\pm}4.63$ & \secondcell{$10.00{\pm}5.57$} & \thirdcell{$52.31{\pm}2.15$} \\
MERCI-prefix & $90.83{\pm}0.80$ & \thirdcell{$60.80{\pm}2.19$} & $20.96{\pm}2.47$ & $52.53{\pm}3.56$ & $64.80{\pm}1.38$ & $0.00{\pm}0.00$ & $3.33{\pm}3.33$ & $51.20{\pm}0.00$ \\
GRPO-prefix & \thirdcell{$93.48{\pm}0.68$} & $57.60{\pm}2.21$ & $21.69{\pm}2.50$ & \thirdcell{$58.08{\pm}3.52$} & \secondcell{$70.64{\pm}1.32$} & \thirdcell{$13.33{\pm}6.31$} & \bestcell{$13.33{\pm}6.31$} & $51.39{\pm}2.15$ \\
GRPO-prefix+IMAX & \bestcell{$94.47{\pm}0.63$} & \bestcell{$69.20{\pm}2.07$} & \thirdcell{$23.90{\pm}2.59$} & \bestcell{$61.11{\pm}3.47$} & \bestcell{$71.06{\pm}1.31$} & \thirdcell{$13.33{\pm}6.31$} & \thirdcell{$6.67{\pm}4.63$} & $51.57{\pm}2.15$ \\
DAPO-prefix & $93.18{\pm}0.69$ & $56.60{\pm}2.22$ & \bestcell{$25.74{\pm}2.66$} & $55.56{\pm}3.54$ & \thirdcell{$69.06{\pm}1.34$} & \bestcell{$20.00{\pm}7.43$} & $0.00{\pm}0.00$ & \bestcell{$56.19{\pm}2.14$} \\
DAPO-prefix+IMAX & \secondcell{$93.78{\pm}0.67$} & \secondcell{$65.80{\pm}2.12$} & \thirdcell{$23.90{\pm}2.59$} & \thirdcell{$58.08{\pm}3.52$} & $64.97{\pm}1.38$ & \thirdcell{$13.33{\pm}6.31$} & \secondcell{$10.00{\pm}5.57$} & \secondcell{$55.08{\pm}0.00$} \\
\midrule
\multicolumn{9}{c}{\textbf{Avg@4}} \\
\midrule
\textbf{Method} & \textbf{GSM8K} & \textbf{MATH} & \textbf{Minerva} & \textbf{GPQA} & \textbf{AMC} & \textbf{AIME24} & \textbf{AIME25} & \textbf{IFEval} \\
\midrule
Base & $59.70{\pm}0.87$ & $30.00{\pm}1.55$ & $10.57{\pm}1.39$ & $24.49{\pm}1.53$ & $22.83{\pm}0.76$ & $3.33{\pm}1.98$ & $1.67{\pm}1.67$ & $26.94{\pm}1.46$ \\
GRPO-full & $86.64{\pm}0.84$ & $59.80{\pm}2.01$ & $20.13{\pm}2.28$ & $25.38{\pm}2.21$ & $9.78{\pm}0.72$ & $7.50{\pm}4.67$ & $3.33{\pm}1.58$ & $29.25{\pm}0.00$ \\
\cmidrule(lr){1-9}
Text Init. & $53.49{\pm}0.84$ & $27.50{\pm}1.47$ & $10.02{\pm}1.25$ & \bestcell{$28.54{\pm}2.22$} & $35.22{\pm}0.96$ & \bestcell{$6.67{\pm}2.92$} & \thirdcell{$0.83{\pm}0.83$} & $28.97{\pm}1.51$ \\
DIVER-prefix & $66.49{\pm}0.87$ & $31.75{\pm}1.49$ & $9.93{\pm}1.27$ & \secondcell{$27.02{\pm}2.05$} & $35.38{\pm}0.96$ & $2.50{\pm}1.84$ & \secondcell{$2.50{\pm}1.39$} & $30.41{\pm}1.53$ \\
MERCI-prefix & $62.51{\pm}0.86$ & $30.75{\pm}1.41$ & $9.56{\pm}1.26$ & $23.11{\pm}1.94$ & $34.92{\pm}0.98$ & $0.00{\pm}0.00$ & \thirdcell{$0.83{\pm}0.83$} & $31.47{\pm}0.00$ \\
GRPO-prefix & $71.97{\pm}0.81$ & \secondcell{$43.45{\pm}1.89$} & \secondcell{$13.33{\pm}1.74$} & \thirdcell{$26.39{\pm}2.07$} & \bestcell{$39.64{\pm}1.00$} & \secondcell{$5.83{\pm}3.10$} & \bestcell{$4.17{\pm}2.10$} & \thirdcell{$31.93{\pm}1.61$} \\
GRPO-prefix+IMAX & \bestcell{$76.14{\pm}0.81$} & \bestcell{$50.45{\pm}1.81$} & \thirdcell{$11.86{\pm}1.46$} & $25.25{\pm}1.89$ & \thirdcell{$38.07{\pm}0.96$} & \thirdcell{$5.00{\pm}2.79$} & \secondcell{$2.50{\pm}1.84$} & $30.64{\pm}1.58$ \\
DAPO-prefix & \thirdcell{$72.02{\pm}0.84$} & $41.65{\pm}1.87$ & \bestcell{$14.52{\pm}1.73$} & $24.37{\pm}1.98$ & \secondcell{$38.32{\pm}1.00$} & \secondcell{$5.83{\pm}2.30$} & $0.00{\pm}0.00$ & \bestcell{$35.95{\pm}1.67$} \\
DAPO-prefix+IMAX & \secondcell{$72.93{\pm}0.82$} & \thirdcell{$42.70{\pm}1.69$} & $10.85{\pm}1.37$ & $23.99{\pm}1.93$ & $35.78{\pm}1.00$ & \secondcell{$5.83{\pm}2.86$} & \secondcell{$2.50{\pm}1.39$} & \secondcell{$33.13{\pm}0.00$} \\
\bottomrule
\end{tabular}
}
\end{table*}

\begin{table*}[t]
\centering
\caption{Full results for Qwen3-8B-Base.}
\label{tab:full_qwen38}
\scriptsize
\resizebox{\textwidth}{!}{
\begin{tabular}{l|cccccccc}
\toprule
\multicolumn{9}{c}{\textbf{Pass@4}} \\
\midrule
\textbf{Method} & \textbf{GSM8K} & \textbf{MATH} & \textbf{Minerva} & \textbf{GPQA} & \textbf{AMC} & \textbf{AIME24} & \textbf{AIME25} & \textbf{IFEval} \\
\midrule
Base & $88.10{\pm}0.89$ & $58.60{\pm}2.20$ & $18.01{\pm}2.33$ & $29.29{\pm}3.24$ & $41.12{\pm}1.42$ & $13.33{\pm}6.31$ & $10.00{\pm}5.57$ & $48.06{\pm}2.15$ \\
GRPO-full & $91.58{\pm}0.76$ & $74.20{\pm}1.96$ & $29.41{\pm}2.77$ & $47.98{\pm}3.56$ & $74.23{\pm}1.26$ & $26.67{\pm}8.21$ & $33.33{\pm}8.75$ & $47.87{\pm}0.00$ \\
\cmidrule(lr){1-9}
Text Init. & $90.30{\pm}0.82$ & $59.80{\pm}2.19$ & $20.96{\pm}2.47$ & \bestcell{$55.56{\pm}3.54$} & $62.39{\pm}1.40$ & \bestcell{$20.00{\pm}7.43$} & \thirdcell{$6.67{\pm}4.63$} & $52.50{\pm}2.15$ \\
DIVER-prefix & $90.75{\pm}0.80$ & $63.00{\pm}2.16$ & \thirdcell{$21.69{\pm}2.50$} & $47.47{\pm}3.56$ & $62.89{\pm}1.40$ & $10.00{\pm}5.57$ & \thirdcell{$6.67{\pm}4.63$} & \bestcell{$55.27{\pm}2.14$} \\
MERCI-prefix & \secondcell{$93.18{\pm}0.69$} & \thirdcell{$66.40{\pm}2.11$} & $18.38{\pm}2.35$ & \bestcell{$55.56{\pm}3.54$} & \secondcell{$66.47{\pm}1.36$} & $10.00{\pm}5.57$ & \thirdcell{$6.67{\pm}4.63$} & \secondcell{$54.71{\pm}2.14$} \\
GRPO-prefix & $91.78{\pm}0.87$ & \thirdcell{$66.40{\pm}2.11$} & $19.12{\pm}2.39$ & \thirdcell{$50.00{\pm}3.56$} & $64.14{\pm}1.39$ & \secondcell{$16.67{\pm}6.92$} & $3.33{\pm}3.33$ & \secondcell{$54.71{\pm}2.14$} \\
GRPO-prefix+IMAX & \bestcell{$94.01{\pm}0.65$} & \bestcell{$67.20{\pm}2.10$} & $20.96{\pm}2.47$ & \bestcell{$55.56{\pm}3.54$} & \bestcell{$67.06{\pm}1.36$} & $10.00{\pm}5.57$ & \secondcell{$10.00{\pm}5.57$} & $51.57{\pm}2.15$ \\
DAPO-prefix & $90.07{\pm}0.82$ & $63.60{\pm}2.15$ & \secondcell{$22.43{\pm}2.53$} & $48.48{\pm}3.56$ & \thirdcell{$65.47{\pm}1.37$} & $10.00{\pm}5.57$ & \thirdcell{$6.67{\pm}4.63$} & \thirdcell{$53.79{\pm}2.15$} \\
DAPO-prefix+IMAX & \thirdcell{$92.72{\pm}0.72$} & \secondcell{$67.00{\pm}2.10$} & \bestcell{$23.16{\pm}2.56$} & \secondcell{$54.04{\pm}3.55$} & $63.55{\pm}1.39$ & \thirdcell{$13.33{\pm}6.31$} & \bestcell{$16.67{\pm}6.92$} & $51.76{\pm}2.15$ \\
\midrule
\multicolumn{9}{c}{\textbf{Avg@4}} \\
\midrule
\textbf{Method} & \textbf{GSM8K} & \textbf{MATH} & \textbf{Minerva} & \textbf{GPQA} & \textbf{AMC} & \textbf{AIME24} & \textbf{AIME25} & \textbf{IFEval} \\
\midrule
Base & $55.44{\pm}0.86$ & $33.65{\pm}1.60$ & $8.73{\pm}1.31$ & $8.59{\pm}1.03$ & $18.58{\pm}0.79$ & $4.17{\pm}2.10$ & $3.33{\pm}1.98$ & $27.26{\pm}1.50$ \\
GRPO-full & $76.95{\pm}0.89$ & $65.30{\pm}1.93$ & $22.33{\pm}2.31$ & $19.57{\pm}1.78$ & $47.46{\pm}1.09$ & $16.67{\pm}5.91$ & $16.67{\pm}5.41$ & $31.24{\pm}0.00$ \\
\cmidrule(lr){1-9}
Text Init. & $56.16{\pm}0.82$ & $33.60{\pm}1.52$ & $8.36{\pm}1.11$ & \bestcell{$22.73{\pm}1.81$} & $31.15{\pm}0.91$ & \bestcell{$5.83{\pm}2.30$} & \thirdcell{$2.50{\pm}1.84$} & $30.87{\pm}1.55$ \\
DIVER-prefix & $57.15{\pm}0.80$ & $36.65{\pm}1.60$ & \thirdcell{$9.56{\pm}1.26$} & $17.05{\pm}1.50$ & $29.69{\pm}0.86$ & \thirdcell{$3.33{\pm}1.98$} & $1.67{\pm}1.16$ & \secondcell{$32.76{\pm}1.58$} \\
MERCI-prefix & $59.59{\pm}0.79$ & $37.95{\pm}1.53$ & $8.64{\pm}1.24$ & \thirdcell{$20.83{\pm}1.68$} & $31.86{\pm}0.87$ & \thirdcell{$3.33{\pm}1.98$} & \thirdcell{$2.50{\pm}1.84$} & \secondcell{$32.76{\pm}1.59$} \\
GRPO-prefix & \secondcell{$60.90{\pm}0.82$} & \thirdcell{$40.20{\pm}1.61$} & $9.19{\pm}1.31$ & $17.93{\pm}1.51$ & \bestcell{$33.36{\pm}0.94$} & \bestcell{$5.83{\pm}2.59$} & $0.83{\pm}0.83$ & \bestcell{$33.60{\pm}1.62$} \\
GRPO-prefix+IMAX & \bestcell{$64.18{\pm}0.79$} & \secondcell{$41.00{\pm}1.64$} & \secondcell{$10.02{\pm}1.35$} & \bestcell{$22.73{\pm}1.84$} & \secondcell{$32.42{\pm}0.87$} & \thirdcell{$3.33{\pm}1.98$} & \secondcell{$3.33{\pm}1.98$} & $30.59{\pm}1.54$ \\
DAPO-prefix & $52.12{\pm}0.79$ & $34.60{\pm}1.48$ & \thirdcell{$9.56{\pm}1.24$} & $18.18{\pm}1.60$ & $31.86{\pm}0.89$ & \thirdcell{$3.33{\pm}1.98$} & \thirdcell{$2.50{\pm}1.84$} & \thirdcell{$31.05{\pm}1.54$} \\
DAPO-prefix+IMAX & \thirdcell{$60.20{\pm}0.77$} & \bestcell{$41.10{\pm}1.64$} & \bestcell{$11.40{\pm}1.47$} & \secondcell{$22.22{\pm}1.84$} & \thirdcell{$32.24{\pm}0.93$} & \secondcell{$4.17{\pm}2.10$} & \bestcell{$6.67{\pm}2.92$} & $30.45{\pm}1.54$ \\
\bottomrule
\end{tabular}
}
\end{table*}


Table~\ref{tab:full_ablation} reports the complete ablation results with standard errors for both Pass@4 and Avg@4. All values are percentages. This table complements Table~\ref{tab:ablation_results} by including the full benchmark set and the one-, two-, and four-prefix variants.

\begin{table*}[t]
\centering
\caption{Full ablation results for prefix-pool training on Qwen2.5-1.5B-Instruct.}
\label{tab:full_ablation}
\scriptsize
\resizebox{\textwidth}{!}{
\begin{tabular}{ll|cccccccc}
\toprule
\textbf{Run} & \textbf{Metric} & \textbf{GSM8K} & \textbf{MATH} & \textbf{Minerva} & \textbf{GPQA} & \textbf{AMC} & \textbf{AIME24} & \textbf{AIME25} & \textbf{IFEval} \\
\midrule
SFT & Pass@4 & $83.40{\pm}1.00$ & $54.80{\pm}2.20$ & $14.70{\pm}2.20$ & $66.20{\pm}3.40$ & $68.80{\pm}1.30$ & $3.30{\pm}3.30$ & $0.00{\pm}0.00$ & $54.70{\pm}2.10$ \\
SFT & Avg@4 & $58.40{\pm}1.00$ & $28.50{\pm}1.40$ & $6.00{\pm}1.00$ & $27.10{\pm}1.80$ & $29.90{\pm}0.80$ & $0.80{\pm}0.80$ & $0.00{\pm}0.00$ & $35.80{\pm}1.70$ \\
IMAX-1 prefix & Pass@4 & $83.60{\pm}1.00$ & $58.60{\pm}2.20$ & $18.80{\pm}2.40$ & $66.20{\pm}3.40$ & $69.30{\pm}1.30$ & $0.00{\pm}0.00$ & $0.00{\pm}0.00$ & $54.50{\pm}2.10$ \\
IMAX-1 prefix & Avg@4 & $64.30{\pm}1.10$ & $40.30{\pm}1.80$ & $9.00{\pm}1.40$ & $27.30{\pm}1.90$ & $34.90{\pm}0.90$ & $0.00{\pm}0.00$ & $0.00{\pm}0.00$ & $39.10{\pm}1.80$ \\
IMAX-2 prefixes & Pass@4 & $85.75{\pm}1.17$ & $60.00{\pm}2.19$ & $18.38{\pm}2.35$ & $62.12{\pm}3.46$ & $68.06{\pm}1.35$ & $0.00{\pm}0.00$ & $3.33{\pm}3.33$ & $53.05{\pm}2.15$ \\
IMAX-2 prefixes & Avg@4 & $67.40{\pm}1.22$ & $42.30{\pm}1.86$ & $9.47{\pm}1.43$ & $26.77{\pm}1.97$ & $34.45{\pm}0.95$ & $0.00{\pm}0.00$ & $0.83{\pm}0.83$ & $36.78{\pm}1.75$ \\
IMAX-4 prefixes & Pass@4 & $86.60{\pm}0.90$ & $58.40{\pm}2.20$ & $19.50{\pm}2.40$ & $61.10{\pm}3.50$ & $67.60{\pm}1.40$ & $6.70{\pm}4.60$ & $3.30{\pm}3.30$ & $54.70{\pm}2.10$ \\
IMAX-4 prefixes & Avg@4 & $68.20{\pm}1.00$ & $41.60{\pm}1.90$ & $9.50{\pm}1.40$ & $26.90{\pm}2.00$ & $34.90{\pm}1.00$ & $1.70{\pm}1.20$ & $0.80{\pm}0.80$ & $37.60{\pm}1.70$ \\
\bottomrule
\end{tabular}
}
\end{table*}

\subsection{Further Analysis}

\begin{figure*}
\centering
\begin{subfigure}[t]{\textwidth}
    \centering
    \includegraphics[width=\linewidth]{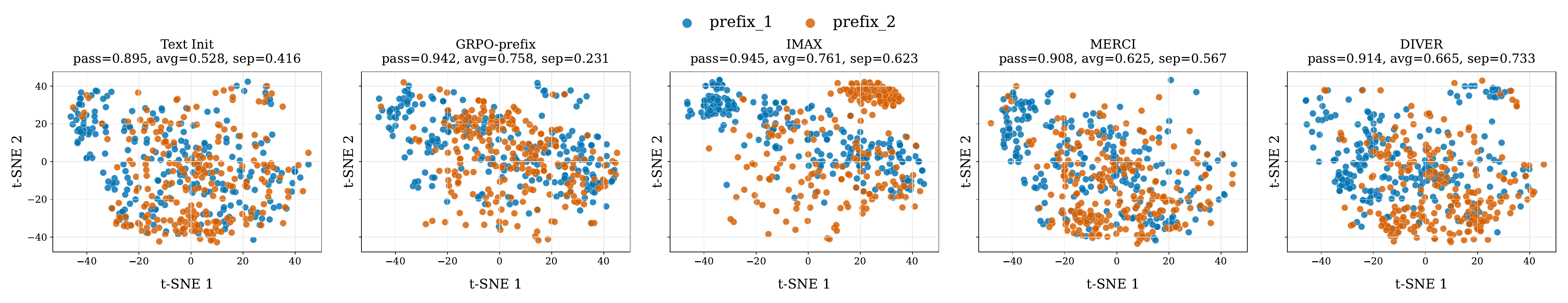}
    \caption{GSM8K.}
    \label{fig:complete_embedding_gsm8k}
\end{subfigure}\hfill
\begin{subfigure}[t]{\textwidth}
    \centering
    \includegraphics[width=\linewidth]{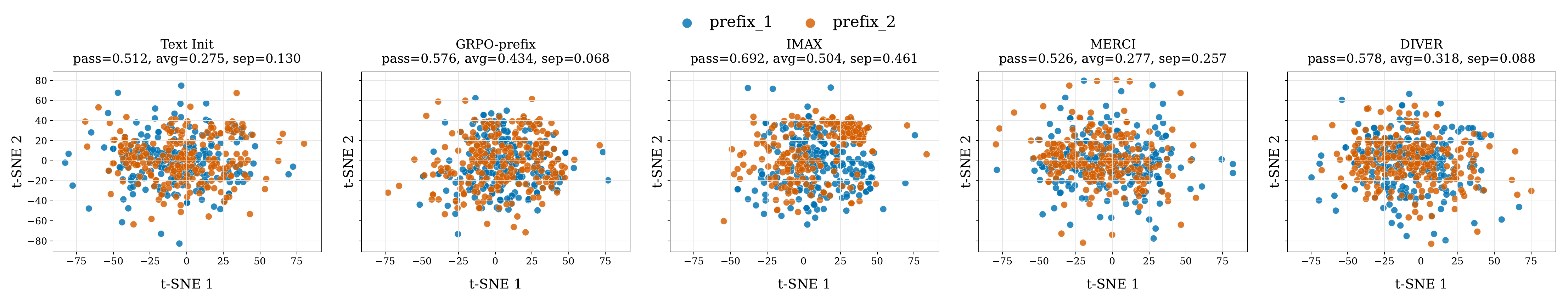}
    \caption{MATH-500.}
    \label{fig:complete_embedding_math500}
\end{subfigure}\hfill
\begin{subfigure}[t]{\textwidth}
    \centering
    \includegraphics[width=\linewidth]{imgs/minerva_math_answer_embedding_space_tsne_minerva_qwen3_initial_grpo_diverse_merci_diver_centered_correct_only.pdf}
    \caption{Minerva Math.}
    \label{fig:complete_embedding_minerva}
\end{subfigure}
\caption{Complete t-SNE visualizations of response embeddings for the two trained prefixes across GSM8K, MATH-500, and Minerva Math. }
\label{fig:complete_embedding_analysis}
\end{figure*}

\begin{figure}
    \centering
    \includegraphics[width=\linewidth]{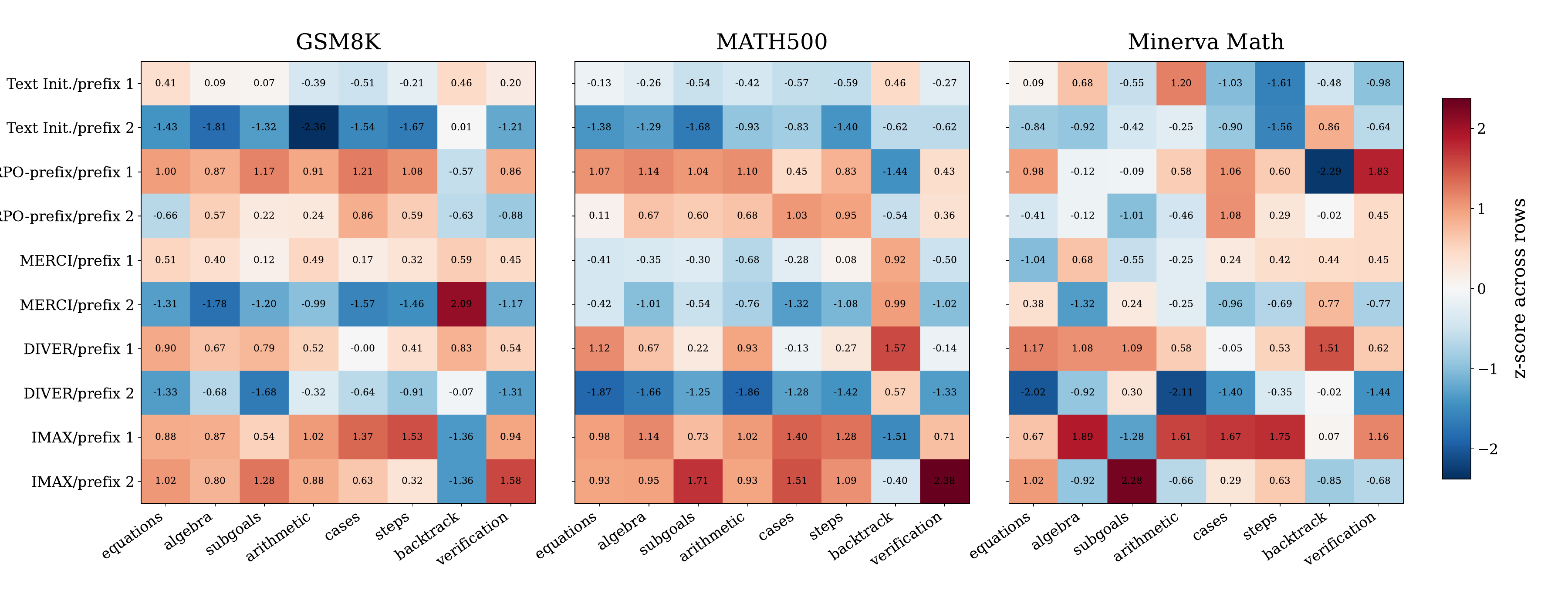}
    \caption{Reasoning-component contrast between Prefix 1 and Prefix 2 on GSM8K, MATH-500 and Minerva MATH.}
    \label{fig:complete_heatmap_analysis}
\end{figure}

\begin{figure}
    \centering
    \includegraphics[width=\linewidth]{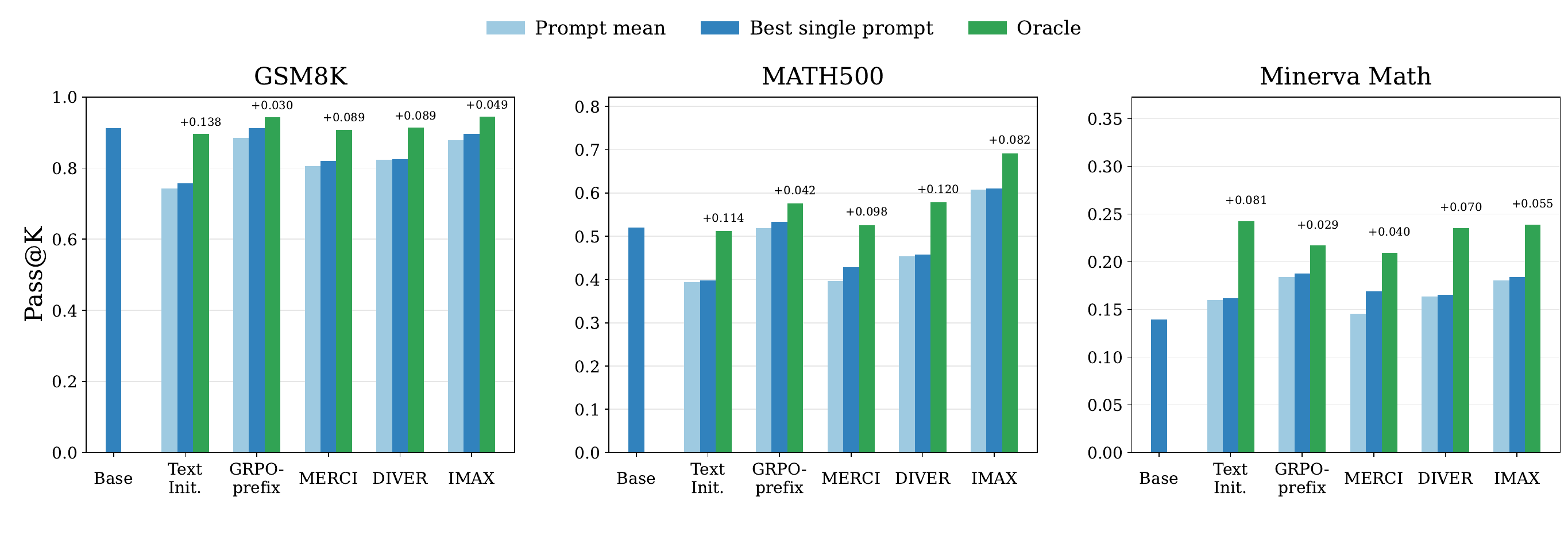}
    \caption{Complementary correctness analysis for the two prefixes. The figure separates examples solved by only one prefix from those solved by both prefixes, illustrating whether prefix tuning improves quality while preserving non-identical successful regions.}
    \label{fig:complete_correctness_analysis}
\end{figure}

\begin{figure*}[t]
\centering
\begin{subfigure}[t]{0.48\textwidth}
    \centering
    \includegraphics[width=\linewidth]{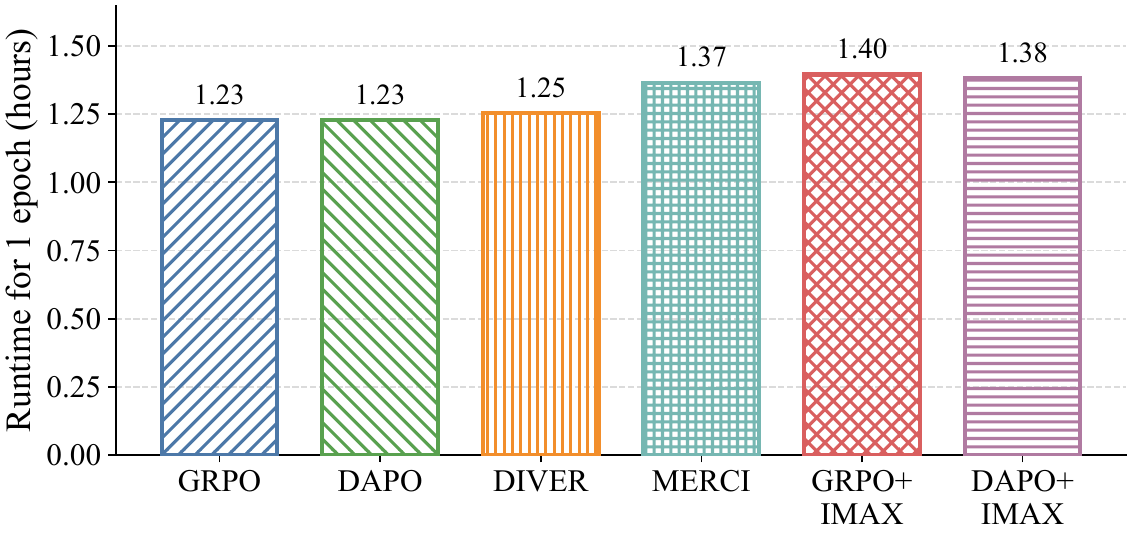}
    \caption{Qwen2.5-1.5B.}
    \label{fig:runtime_qwen25}
\end{subfigure}\hfill
\begin{subfigure}[t]{0.48\textwidth}
    \centering
    \includegraphics[width=\linewidth]{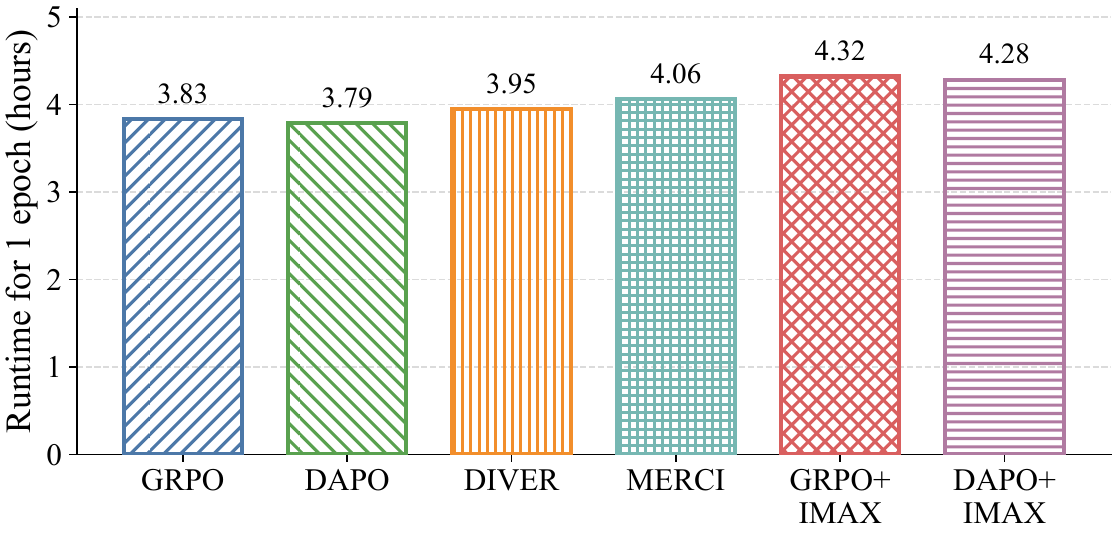}
    \caption{Qwen3-4B.}
    \label{fig:runtime_qwen34}
\end{subfigure}

\vspace{0.6em}
\begin{subfigure}[t]{0.48\textwidth}
    \centering
    \includegraphics[width=\linewidth]{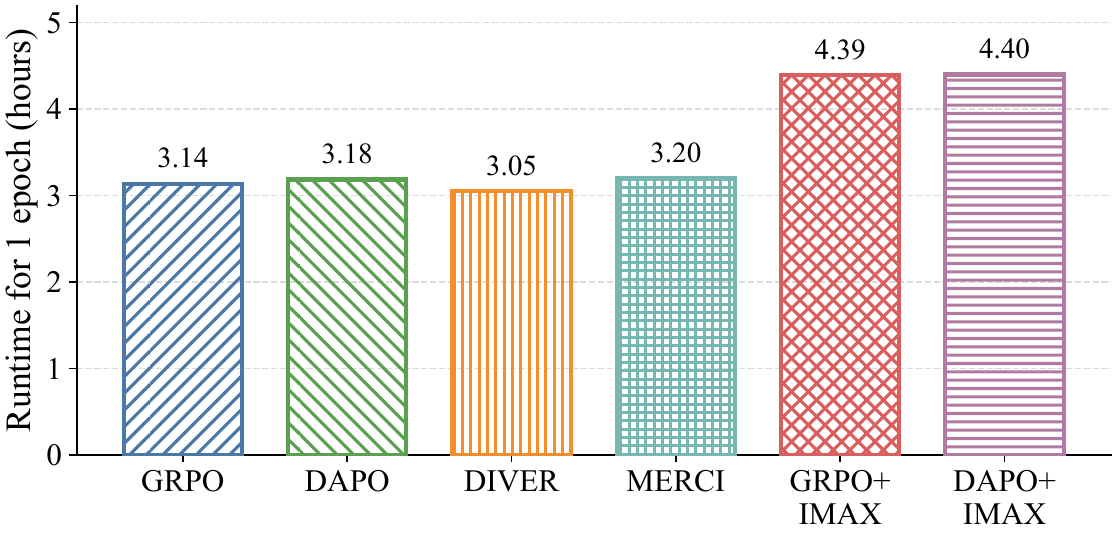}
    \caption{Qwen3-8B.}
    \label{fig:runtime_qwen38}
\end{subfigure}\hfill
\begin{subfigure}[t]{0.48\textwidth}
    \centering
    \includegraphics[width=\linewidth]{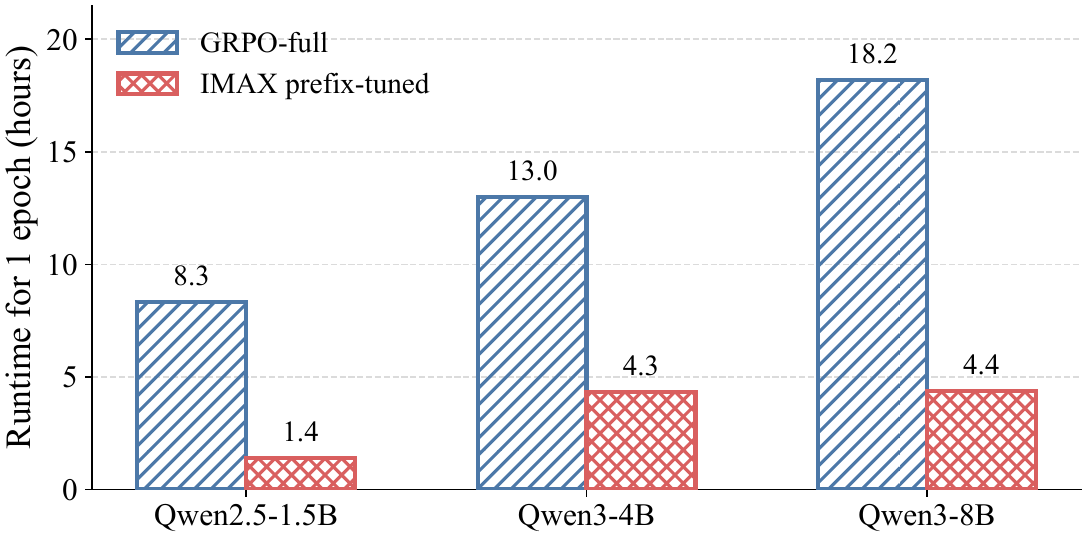}
    \caption{GPU-hour comparison with GRPO-full.}
    \label{fig:runtime_gpu_hours}
\end{subfigure}
\caption{Runtime analysis for prefix-tuned methods and full-parameter GRPO. The first three panels report one-epoch runtime for each method under the prefix-tuning setting. The last panel compares IMAX with full-parameter GRPO using GPU-hours to account for the different number of GPUs used in training.}
\label{fig:runtime_analysis}
\end{figure*}

We provide additional figures that support the qualitative conclusions in Section~\ref{sec:analysis}. Figure~\ref{fig:complete_embedding_analysis} shows response-embedding structure across three reasoning benchmarks, Figure~\ref{fig:complete_heatmap_analysis} summarizes reasoning-component differences between prefixes, Figure~\ref{fig:complete_correctness_analysis} compares complementary correctness across prefixes, and Figure~\ref{fig:runtime_analysis} reports runtime cost.

For the embedding figures, we report a normalized centroid-separation score between \texttt{pool\_001} and \texttt{pool\_002} in the plotted two-dimensional t-SNE space. Let $\mathcal{X}_1=\{x_i:\text{point }i\text{ uses }\texttt{pool\_001}\}$ and $\mathcal{X}_2=\{x_i:\text{point }i\text{ uses }\texttt{pool\_002}\}$, where $x_i\in\mathbb{R}^2$ is the plotted coordinate. We compute
\begin{equation}
\mu_k=\frac{1}{|\mathcal{X}_k|}\sum_{x\in\mathcal{X}_k}x,\quad
s_k=\frac{1}{|\mathcal{X}_k|}\sum_{x\in\mathcal{X}_k}\|x-\mu_k\|_2,\quad
\mathrm{sep}=\frac{\|\mu_1-\mu_2\|_2}{\max\left((s_1+s_2)/2,10^{-12}\right)}.
\end{equation}
Higher separation therefore means that the two prefix clusters are farther apart relative to their average within-prefix spread. The plotted points are correct-only responses and are example-centered before t-SNE projection.

\section{Detailed Experiment Setup}
\label{app:experiment_setup}

\subsection{Implementation Details}

We train only the soft-prefix pool while keeping the backbone LLM frozen. For each prompt, the training procedure assigns prefixes in a stratified manner and samples rollouts under the corresponding prefix-conditioned distribution. For IMAX, the posterior model $q_\phi(z\mid x,y)$ is trained from prompt--response pairs using the frozen backbone representation and a trainable classification head; the resulting response-level InfoMax reward is then added to the verifiable reward before the prefix update. The main training and decoding hyperparameters are summarized in Table~\ref{tab:training_settings}.

\subsection{Default Prefix Pool}
\label{app:default_prefix_pool}

The default text prompt pool used to initialize the soft-prefix pool is shown in Table~\ref{tab:default_prefix_pool}.

\begin{table*}[htbp]
\centering
\caption{Default text prompt pool for soft-prefix initialization.}
\label{tab:default_prefix_pool}
\footnotesize
\setlength{\tabcolsep}{5pt}
\renewcommand{\arraystretch}{1.18}
\begin{tabular}{>{\centering\arraybackslash}p{0.06\textwidth}|p{0.87\textwidth}}
\toprule
\textbf{ID} & \textbf{Text Prompt} \\
\midrule
1 & Reasoning Strategy: Introduce variables explicitly, write down the governing equations or identities, and solve through exact symbolic manipulation. Prefer precise derivation over heuristic guessing. \\
2 & Reasoning Strategy: Partition the problem into exhaustive, mutually exclusive cases. Solve each case carefully, rule out impossible ones, and combine the valid conclusions. \\
3 & Reasoning Strategy: Transform the problem into an equivalent but simpler form. Use substitution, reparameterization, factorization, or an alternative representation if it exposes the structure. \\
4 & Reasoning Strategy: Look for an invariant, symmetry, conserved quantity, or monotonic property. Re-express the problem around that structure before solving. \\
5 & Reasoning Strategy: Examine small, boundary, or special cases to discover a pattern or conjecture. Then generalize it and verify it rigorously for the full problem. \\
6 & Reasoning Strategy: List the strongest constraints first and use them to eliminate impossible values, configurations, or structures before detailed computation. \\
7 & Reasoning Strategy: Construct the required object, value, or configuration explicitly when possible, and then prove that the construction satisfies all conditions. \\
8 & Reasoning Strategy: Focus on an extreme case, maximal or minimal element, or assume the opposite of the target claim and derive a contradiction to force the solution. \\
\bottomrule
\end{tabular}
\end{table*}

\begin{table*}[htbp]
    \centering
    \caption{Training and decoding settings used for prefix-tuned RLVR experiments.}
    \label{tab:training_settings}
    \small
    \setlength{\tabcolsep}{10pt}
    \begin{tabular}{l|c}
        \toprule
        \textbf{Settings} & \textbf{Value} \\
        \midrule
        Trainable parameters      & Soft prefixes and $q_\phi$ head \\
        Backbone LLM              & Frozen \\
        Num virtual tokens        & 32 \\
        Max steps                 & 300 \\
        Max response length       & 2048 \\
        Learning rate             & $5{\times}10^{-4}$ \\
        $q_\phi$ learning rate    & $1{\times}10^{-4}$ \\
        Intrinsic reward weight   & 0.01 \\
        Per-device train batch size & 4 \\
        Per-device eval batch size & 8 \\
        Gradient accumulation steps & 8 \\
        Training temperature      & 1.0 \\
        Prefix assignment         & Stratified \\
        Number of prefixes $C$    & 2 \\
        Group size per prompt--prefix pair $N$ & 2 \\
        Evaluation rollouts $N_\mathrm{eval}$ & 4 \\
        Evaluation temperature    & 0.7 \\
        Evaluation top-$k$        & 20 \\
        Evaluation top-$p$        & 0.8 \\
        \bottomrule
    \end{tabular}
\end{table*}

For evaluation, we use stratified prefix assignment and sample $N_\mathrm{eval}=4$ responses per problem unless otherwise specified. We report Pass@4 and Avg@4 on reasoning benchmarks and IFEval, matching the metrics used in the main text. The full result tables in Appendix~\ref{app:additional_results} include standard errors for both metrics.


\section{Algorithm}
\label{app:algorithm}

Algorithm~\ref{alg:imax} summarizes the online training procedure of IMAX. The key implementation detail is that the posterior model $q_\phi$ is updated from the original prompt--response pair without the soft prefix, while the RLVR update is applied only to the soft-prefix pool under the frozen backbone.

\begin{algorithm}[htbp]
   \caption{IMAX}
   \label{alg:imax}
\begin{algorithmic}
   \STATE {\bfseries Input:} base RLVR algorithm $\mathcal{A}$, prompt distribution $p(x)$, prefix prior $p(z)$ over $C$ prefixes, group size $N$, total steps $T$, InfoMax reward weight $\beta$
   \STATE {\bfseries Initialize:} prefix pool $\mathcal{E}$ from text prompts, posterior head parameters $\phi$

   \FOR{$t=1$ {\bfseries to} $T$}
      \STATE Sample prompts $\{x_i\}_{i=1}^{B} \sim p(x)$
      \STATE For each $x_i$, assign distinct prefix identities $\{z_{i,c}\}_{c=1}^{C}$
      \STATE Collect rollouts $y_{i,c}^{(n)} \sim \pi_{\mathcal{E}}(\cdot \mid x_i,z_{i,c})$, $n=1,\ldots,N$
      \STATE Compute verifiable rewards $r_{i,c}^{(n)}$ from rollouts

      \STATE Fix $\mathcal{E}$ and update $q_\phi$ to predict $z_{i,c}$ from each prompt--response pair $(x_i,y_{i,c}^{(n)})$ without the soft prefix
      \STATE Compute response-level InfoMax rewards $r^{\mathrm{InfoMax}}_{i,c,n} \leftarrow \log q_\phi(z_{i,c}\mid x_i,y_{i,c}^{(n)})$
      \STATE Form augmented rewards $\tilde r_{i,c}^{(n)} \leftarrow r_{i,c}^{(n)} + \beta\, r^{\mathrm{InfoMax}}_{i,c,n}$
      \STATE Fix $q_\phi$ and update $\mathcal{E}$ using $\mathcal{A}$ with augmented rewards $\{\tilde r_{i,c}^{(n)}\}$
   \ENDFOR
   \STATE {\bfseries Output:} trained prefix pool $\mathcal{E}$
\end{algorithmic}
\end{algorithm}

\end{document}